\documentclass[final,5p,times,twocolumn,authoryear]{elsarticle}

\usepackage{amsmath,amssymb,amsfonts}
\usepackage[utf8]{inputenc} % allow utf-8 input
\usepackage[T1]{fontenc}    % use 8-bit T1 fonts

\usepackage{natbib}
\usepackage[pagebackref=true,breaklinks=true,colorlinks,bookmarks=false]{hyperref}

\usepackage{graphicx}

\usepackage{url}
\usepackage{bbm}
\usepackage{makecell}

\usepackage{booktabs}
\usepackage{multirow, multicol}
\usepackage{caption}
\usepackage{subcaption}

\usepackage{microtype}
\usepackage{nicefrac}

\usepackage{color,soul}
\usepackage[dvipsnames]{xcolor}

\newcommand{\yh}{\hat{Y}}
\newcommand{\separator}{\;|\;}
\DeclareMathOperator{\acc}{acc}
\DeclareMathOperator{\conf}{conf}
\journal{Computer Methods and Programs in Biomedicine}
\date{}

\begin{document}

\begin{frontmatter}

%% Title, authors and addresses

%% use the tnoteref command within \title for footnotes;
%% use the tnotetext command for theassociated footnote;
%% use the fnref command within \author or \affiliation for footnotes;
%% use the fntext command for theassociated footnote;
%% use the corref command within \author for corresponding author footnotes;
%% use the cortext command for theassociated footnote;
%% use the ead command for the email address,
%% and the form \ead[url] for the home page:
%% \title{Title\tnoteref{label1}}
%% \tnotetext[label1]{}
%% \author{Name\corref{cor1}\fnref{label2}}
%% \ead{email address}
%% \ead[url]{home page}
%% \fntext[label2]{}
%% \cortext[cor1]{}
%% \affiliation{organization={},
%%            addressline={}, 
%%            city={},
%%            postcode={}, 
%%            state={},
%%            country={}}
%% \fntext[label3]{}

\title{Understanding Calibration of Deep Neural Networks for Medical Image Classification}

%% use optional labels to link authors explicitly to addresses:
\author[label1]{Abhishek Singh Sambyal}
\ead{abhishek.19csz0001@iitrpr.ac.in}

\author[label1]{Usma Niyaz}
\ead{usma.20csz0015@iitrpr.ac.in}

\author[label2]{Narayanan C. Krishnan \corref{cor2}}
\ead{ckn@iitpkd.ac.in}
\cortext[cor2]{He is on lien from the Indian Institute of Technology Ropar.}

\author[label1]{Deepti R. Bathula}
\ead{bathula@iitrpr.ac.in}

\affiliation[label1]{organization={Department of Computer Science and Engineering, Indian Institute of Technology Ropar},
            city={Rupnagar},
            postcode={140001},
            state={Punjab},
            country={India}}

\affiliation[label2]{organization={Department of Data Science, Indian Institute of Technology Palakkad},
            city={Palakkad},
            postcode={678532},
            state={Kerala},
            country={India}}

\begin{abstract}
\textbf{Background and Objective \textendash} In the field of medical image analysis, achieving high accuracy is not enough; ensuring well-calibrated predictions is also crucial. Confidence scores of a deep neural network play a pivotal role in explainability by providing insights into the model's certainty, identifying cases that require attention, and establishing trust in its predictions. Consequently, the significance of a well-calibrated model becomes paramount in the medical imaging domain, where accurate and reliable predictions are of utmost importance. While there has been a significant effort towards training modern deep neural networks to achieve high accuracy on medical imaging tasks, model calibration and factors that affect it remain under-explored.\\[5pt]
\textbf{Methods \textendash} To address this, we conducted a comprehensive empirical study that explores model performance and calibration under different training regimes. We considered fully supervised training, which is the prevailing approach in the community, as well as rotation-based self-supervised method with and without transfer learning, across various datasets and architecture sizes. Multiple calibration metrics were employed to gain a holistic understanding of model calibration.\\[5pt]
\textbf{Results \textendash} Our study reveals that factors such as weight distributions and the similarity of learned representations correlate with the calibration trends observed in the models. Notably, models trained using rotation-based self-supervised pretrained regime exhibit significantly better calibration while achieving comparable or even superior performance compared to fully supervised models across different medical imaging datasets.\\[5pt]
\textbf{Conclusion \textendash} These findings shed light on the importance of model calibration in medical image analysis and highlight the benefits of incorporating self-supervised learning approach to improve both performance and calibration.
\end{abstract}

\begin{keyword}
Calibration \sep deep neural network \sep fully-supervised \sep self-supervised \sep transfer learning \sep medical imaging.
\end{keyword}

\end{frontmatter}

% \linenumbers

%% main text
\section{Introduction}
\label{sec:introduction}
\noindent Recent advances in deep neural networks have shown remarkable improvement in performance for many computer vision tasks like classification, segmentation, and object detection \citep{pretraining-krizhevsky, mask-cnn}. However, it is essential that model predictions are not only accurate but also well calibrated \citep{ece-guo}. Model calibration refers to the accurate estimation of the probability of correctness or uncertainty of its predictions. As calibration directly relates to the trustworthiness of a model's predictions, it is an essential factor for evaluating models in safety-critical applications like medical image analysis \citep{clinical-jiang, calibration-second-opinion, trustworthy-scaricity-medicalimaging, trustworthiness}.

Probabilities derived from deep learning models are often used as the basis for interpretation because they provide a measure of confidence or certainty associated with the predictions. When a deep learning model assigns a high probability to a particular class, it indicates a stronger belief in that prediction. For example, in medical diagnosis, a high probability assigned to a certain disease can indicate a higher likelihood of its presence based on the observed input data. However, it is important to note that the reliability of interpretation based on probabilities depends on the calibration of the model \citep{confidence-reliability, ece-guo, caruana}. Calibration ensures that the assigned probabilities reflect the true likelihood of events, allowing for accurate interpretation. Without proper calibration, the interpretation based solely on probabilities may be misleading or unreliable.

Apart from directly interpreting the probabilities as confidence for decision process, several explainability methods \citep{xai-medical-survey} have been proposed that depend on the information extracted from the model predictions like weighting random masks \citep{xai-rise}, perturbation \citep{xai-mp, xai-mp-medical}, prediction difference analysis \citep{xai-pda}, contribution scores \citep{xai-deeplift}. The contribution of calibration to the model's explainability lies in providing reliable probability estimates, which aid in understanding the model's decision-making process and associated uncertainties. It is observed that the improved calibration has a positive impact on the saliency maps obtained as interpretations, also improving their quality in terms of faithfulness and are more human-friendly \citep{calibrate-intrepret}. This interplay between explainability and calibrated predictions emerges as a pivotal factor in establishing a trustworthy model for medical decision support systems.

In healthcare, even minor errors in model prediction can carry life-threatening consequences. Therefore, incorporating uncertainty assessment into model predictions can lead to more principled decision-making that safeguards patient well-being. For example, human expertise can be sought in cases with high uncertainty. A model's predictive uncertainty is influenced by noise in data, incomplete coverage of the domain, and imperfect models. Effectively estimating or minimizing these uncertainties can markedly enhance the overall quality and reliability of the results \citep{uncertainty-subject-dataset-level-reyes, uncertainty-reliability-reyes}. Considerable endeavors have been dedicated to mitigating both data and model uncertainty through strategies like data augmentation \citep{sambyal-aleatoric-isbi, uncertainty-aleatoric-tta}, Bayesian inference \citep{bnn, mc-dropout, bnn-awate}, and ensembling \citep{uncertainty-segmentation-mehrtash, calibration-lakshminarayanan}

Modern neural networks are known to be miscalibrated \citep{ece-guo} (overconfident, i.e., high confidence but low accuracy, or underconfident, i.e., low confidence but high accuracy). Hence, model calibration has drawn significant attention in recent years. Approaches to improve the calibration of deep neural networks include post-hoc strategies \citep{calibration-platt, ece-guo}, data augmentation \citep{mixup, overconfidenceerror-mixup, calibration-augmix-hendrycks} and ensembling \citep{calibration-lakshminarayanan}. Similar strategies have also been utilized in the domain of medical image analysis to explore calibration with the primary goal of alleviating miscalibration \citep{weightscaling-miccai, meep-miccai, devil-margin-media, domino-miccai}. Furthermore, recent research has also investigated the impact of different training approaches on the model's performance and calibration. These include the use of focal loss \citep{focalloss}, self-supervised learning \citep{ssl-hendrycks}, and fully-supervised networks with pretraining \citep{rmsce-hendrycks}. However, the scope of these studies has been limited to exploring calibration in the context of generic computer vision datasets like CIFAR10, CIFAR100, and ImageNet \citep{ssl_transfer, calibration-study}. Moreover, the majority of these studies have only utilized Expected Calibration Error (ECE) as the calibration metric. Unfortunately, ECE has several drawbacks, rendering it unfit for tasks like multi-class classification and inefficient due to bias-variance trade-off \citep{calibration-metrics}. Nevertheless, as reliable and accurate estimation of predictive uncertainty is important, measuring calibration is an ongoing active research area resulting in many new metrics \citep{calibration-metrics, calibration-dark, overconfidenceerror-mixup, ece-guo, emnlp_calibration}.

Model calibration is tied to the training process that is  inherently challenging for medical image analysis applications. The scarcity of labeled training datasets is a major cause for concern \citep{roadmap-ai-medical-imaging, uncertainty_ensembles}. Gathering labeled data for the medical domain is a daunting task due to the complex and intricate annotating process requiring domain expertise. \textit{Transfer learning} is a popular learning paradigm to circumvent the labeled training data scarcity \citep{radimagenet, trustworthy-scaricity-medicalimaging}. Although transfer learning improves model accuracy, especially for smaller datasets, it also improves the quality of various complementary model components like adversarial robustness, and uncertainty \citep{rmsce-hendrycks}. Remarkably, the literature suggests that the advantages of popular methods such as transfer learning on classical computer vision datasets do not extend to medical imaging applications \citep{transfusion}. \textit{Self-supervised learning (SSL)} is another promising training regime when learning from scarce labeled data in classical computer vision applications \citep{sslgeneralizable2021, ssl-github}. Though fully-supervised (pretrained) and self-supervised approaches seem to improve various model performance measures like accuracy, robustness, and uncertainty \citep{ssl-hendrycks, ssl_robustness}, the impact of the training regime(s) on model calibration is under-explored.

Our current work addresses these crucial gaps in the literature – understanding the calibration of deep neural networks for medical image analysis in the context of different training regimes and several calibration metrics. Accordingly, our main contributions are:
\begin{enumerate}
    \item We study the effect of different training regimes on the performance and calibration of models used for medical image analysis. Specifically, we compare three different training paradigms: Fully-Supervised with random initialization ($FS_r$), Fully-Supervised with pretraining ($FS_p$), and Rotation-based Self-Supervision with pretraining ($SSL_p$).
    \item {We leverage several complementary calibration metrics to provide an accurate, unbiased, and comprehensive evaluation of the predictive uncertainty of models.}
    \item We assess the influence of varying dataset sizes, architecture capacities, and task complexity on the performance and calibration of the models. 
    \item We identified some of the potential factors that are correlated with the observed changes in the calibration of  models. These include layer-wise learned representations as well as the weight distribution of the model parameters.
\end{enumerate}

In general, we observe that the rotation-based self-supervised pretrained training approach provides better calibration for medical image analysis tasks than its fully supervised counterpart, with on-par or better performance. Additionally, our findings contradict recent literature \citep{transfusion} that remarked \textit{``transfer offers little benefit to performance”} for medical datasets. Furthermore, both the weight distribution and the learned representation analysis indicate that self-supervised training provides implicit regularization that in-turn achieves better calibration.

\section{Methods}
\label{sec:methods}
% \section{Training Regimes and Metrics}
% \label{sec:modelsmetrics}
\subsection{Training Regimes}
\label{subsec:trainingregimes}
\subsubsection{Fully-Supervised and Transfer Learning}
\label{subsubsec:fsmodels}
\noindent In a fully-supervised training regime, we use the given input data and the corresponding target value to learn the task. We can train models using two different ways, learning from scratch, i.e., initializing model weights randomly, or pretraining, i.e., transferring knowledge from one task to another by using the learned weights. In the transfer learning approach, a model is first pretrained using supervised learning on a large labeled dataset \citep{pretraining-krizhevsky, pretraining-decaf}. Then the learned generic representations are fine-tuned on the in-domain medical data \citep{transfusion, pretraining-medicalimaging}. Generally, fine-tuning a pretrained model achieves better generalized performance and faster convergence than training a fully-supervised network from scratch \citep{google-medical-imaging, pretraining-rcnn}.\\
We have considered $FS_r$ as a baseline in our experiments where the model is trained from scratch. ImageNet pretraining is used as the default pretraining approach, which has shown remarkable performance on medical imaging datasets \citep{pretraining-medicalimaging}.

\subsubsection{Self-Supervised Learning}
% \begin{figure}[htbp] 
\begin{figure}[t] 
\centerline{
\includegraphics[width=\linewidth,keepaspectratio]{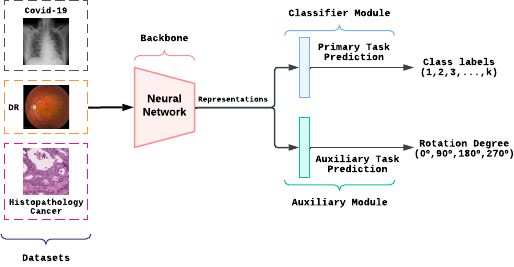}}
\caption{Self-Supervised Learning Framework}
\label{fig:sslblockdiagram}
\end{figure}
\noindent In self-supervised training regime \citep{ssl-hendrycks, rotationnet}, Figure \ref{fig:sslblockdiagram}, we train a classifier network with a separate auxiliary head to predict the induced rotation in the image. The output of the penultimate layer is given to both the classifier and the auxiliary module. The classifier predicts a k-way softmax output vector based on the chosen task/dataset, whereas the auxiliary module predicts a 4-way softmax output vector indicating the rotation degree (0\textdegree, 90\textdegree, 180\textdegree and 270\textdegree). 
Given a dataset $\mathcal{D}$, of $N$ training examples, $\mathcal{D} = \{{x}_i, y_i\}_{i=1}^N$, the goal is to learn representations using a self-supervised regime. The overall loss during training is the weighted sum of vanilla classification and the auxiliary task loss
\begin{equation}
\label{eqn:sslloss}
\begin{split}
\mathcal{L}(\theta) = \mathcal{L}(y, p(y|R_r(x));\theta) + \lambda \mathcal{L}_{aux}(r, p(r|R_r(x));\theta)
\end{split}
\end{equation}
where, $R_r(x)$ is a rotation transformation on input image $x$ and $r \in \{0^{\circ}, 90^{\circ}, 180^{\circ}, 270^{\circ}\}$ is the ground truth label for the auxiliary task. Note that the auxiliary component does not require ground truth training label $y$ as input. $\mathcal{L}_{aux}$ is the cross-entropy between $r$ and the predicted rotation.

\subsection{Calibration Metrics}
\label{sec:calibrationmetrics}
\noindent
\textit{Perfect Calibration}: In a multi-class classification problem, let the input be $X$ and the label $Y \in \{1, 2, \cdots, K\}$ and $f$ the learned model. The model's output is $f(X)=(\hat{Y}, \hat{P})$ where $\hat{Y}$ is a class prediction and $\hat{P}$ is its associated confidence. If $\hat{P}$ is always the true probability, then we call the model perfectly calibrated as defined in \eqref{eqn:perfectcalibration}.
\begin{align}
  \label{eqn:perfectcalibration}
  \mathbb{P}\left(\yh = Y \separator \hat{P} = p\right) = p, \quad \forall p \in [0, 1]
\end{align}
The difference between the true confidence (accuracy) and the predicted confidence (output probability), $|\mathbb{P}\left(\yh = Y \separator \hat{P} = p\right) - p|$ for a given $p$ is known as calibration error or miscalibration. Note that $\hat{P}$ is a continuous random variable, the probability in \eqref{eqn:perfectcalibration} cannot be computed using finitely many samples resulting in different approximations for the calibration error as discussed below.

\subsubsection{Expected Calibration Error (ECE)}
\label{subsec:ece}
\noindent The most common miscalibration measure is the ECE \citep{ece-naeini, ece-guo}, which computes the difference in the expectation between confidence and accuracy. It is a scalar summary statistic of calibration.
\begin{align}
  \label{eqn:expectationece}
  \mathbb{E}_{\hat{P}}\left[\left|\mathbb{P}\left(\yh = Y \separator \hat{P} = p\right) - p\right|\right]
\end{align}
In practice, we cannot estimate ECE without quantization; therefore, the confidence scores for the predicted class are divided into \emph{m} equally spaced bins. For each bin, the average confidence ($\conf$) and accuracy ($\acc$) are computed. The difference between the average confidence and accuracy weighted by the number of samples summed over the bins gives us the ECE measure. Formally,
\begin{equation}
\label{eqn:ece}
\text{ECE} = \sum_{m=1}^{M}\frac{n_m}{N}|\acc(m) - \conf(m)|
\end{equation}
where $n_m$ is the number of predictions in bin $m$. While ECE is used extensively to measure calibration, it has some major drawbacks \citep{calibration-metrics}:
\begin{enumerate}
    \item[(i)] Structured around binary classification, ECE only considers the class with maximum predicted probability. As a result, it discounts the accuracy with which the model predicts other class probabilities in a multi-class classification setting.
    \item[(ii)] Deep neural network predictions are typically overconfident, causing skewness in the output probabilities. Consequently, equal-interval binning metrics like ECE is impacted by only a few bins.
    \item[(iii)] The number of bins, as a hyperparameter, plays a crucial role in the quality of calibration estimation. However, determining the optimal number of bins is challenging due to the bias-variance tradeoff.
    \item[(iv)] In a static binning scheme like ECE, overconfident and underconfident predictions occurring in the same bin result in a reduction of calibration error. In such cases, it is difficult to infer the true cause of improvement in model calibration.
\end{enumerate}
These issues have resulted in the development of novel calibration metrics discussed in the following subsections.
\subsubsection{Adaptive Calibration Error (ACE)} 
\noindent As \emph{ECE} suffers from skewness in the output predictions, ACE mainly focuses on the regions where the predictions are made. It uses an adaptive binning scheme to ensure an equal number of predictions in each bin \citep{emnlp_calibration, calibration-metrics}. Formally,
\begin{equation}
\label{eqn:ace}
    \operatorname{ACE} = \frac{1}{KR}\sum_{k=1}^K \sum_{r=1}^{R} \left| \operatorname{acc}(r, k) - \operatorname{conf}(r, k) \right|
\end{equation}
where, $\operatorname{acc}(r, k)$ and $\operatorname{conf}(r, k)$ represent the accuracy and confidence for the adaptive calibration range or bin $r$ and class label $k$, respectively. Due to adaptive binning, the bin spacing can be unequal; wide in the areas where the number of data points is less, and narrow otherwise.

\subsubsection{Maximum Calibration Error (MCE)} 
\noindent It refers to the upper-bound estimate of miscalibration useful in safety-critical applications. MCE \citep{ece-naeini, ece-guo} captures the worst-case deviation between confidence and accuracy by measuring the maximum difference across all bins $m$, as shown below:
\begin{equation}
\label{eqn:mce}
\operatorname{MCE} = \max_{m \in \{1,\ldots,M\} }\left|\acc(m) - \conf(m)\right|
\end{equation}

\subsubsection{Overconfidence Error (OE)}
\noindent Modern deep neural networks provide high confident outputs despite being inaccurate. Thus a metric that captures the model's overconfidence provides better model insights. OE \citep{overconfidenceerror-mixup} captures the overconfidence in the model prediction by penalizing the confidence score only when the model confidence is greater than the accuracy.
\begin{equation}
\label{eqn:oe}
\operatorname{OE} = \sum_{m=1}^M \frac{n_m}{N} \left [ \conf(m) \times \max \Big (\conf(m) - \acc(m) ,0 \Big ) \right ]
\end{equation}

\subsubsection{Brier or Quadratic Score}
\noindent It is a strictly proper scoring rule that measures the accuracy of the probabilistic predictions \citep{calibration-brier, brier-proper, calibration-brier-multi}. It is the mean squared difference between one-hot encoded true label and predicted probability. Formally,
\begin{align}
  \label{eqn:brier}
    \operatorname{Brier} = \sum_{k=1}^K (\mathbbm{1}_{[Y = k]} - \hat{P}(Y = k \separator X))^2
\end{align}

\subsubsection{Negative Log Likelihood (NLL)}
\noindent For safety-critical applications, using a probabilistic classifier that predicts the correct class and gives the probability distribution of the target classes is encouraged. Using NLL, we can evaluate models with the best predictive uncertainty by measuring the quality of the probabilistic predictions \citep{calibration-brier-nll, calibration-kull-brier-nll, calibration-nll}.
Formally,
\begin{align}
  \label{eqn:nll}
  \operatorname{NLL} = - \sum_{k=1}^K \mathbbm{1}_{[Y = k]}\log[\hat{P}(Y = k \separator X)]
\end{align}
Additionally, \textit{Root Mean Square Calibration Error (RMSCE)} \citep{emnlp_calibration, rmsce-hendrycks, rmsce-hendrycks2019oe} measures the square root of the expected squared difference between confidence and accuracy. As it defines the magnitude of miscalibration, it is highly correlated to ECE. Similar to ACE, \textit{Static Calibration Error (SCE)} \citep{calibration-metrics}, extends ECE by measuring calibration over all classes in each bin for a multi-class setting but does not use an adaptive binning approach. As a result, we exclude these metrics from our experimental analysis.

\noindent It can be observed from the above definitions that none of the individual metrics takes a holistic approach. Hence, it is important to recognize that individual metrics are limited in their ability to provide accurate estimates of calibration. Consequently, a collective evaluation of these metrics is necessary for a better or unbiased understanding of calibration performance.

\subsection{Experimental Setup}
\label{sec:experimentalsetup}

\subsubsection{Datasets}
\label{subsec:dataset}
\noindent We used three different datasets to investigate the classification performance and calibration of models trained under different regimes. The datasets have varying characteristics such as  different imaging modalities, and sizes.
\begin{itemize}
    \item The Diabetic Retinopathy (DR) dataset contains 35K high-resolution ($\sim5000 \times 3000$) retinal fundus scans \citep{dataset-eyepacs}. Each image is rated for the severity of diabetic retinopathy on a scale of 0-4, which makes it a five-class classification problem.
    The images are captured under varying imaging conditions, like different models and camera types.
    \item The Histopathologic Cancer dataset contains 220K images (patches of size $96 \times 96$) extracted from larger digital pathological scans \citep{dataset-histopathology2, dataset-histopathology1, dataset-histopathology3}. Each image is annotated with a binary label indicating the presence of tumor tissue in the histopathologic scans of lymph node sections.
    \item The COVID-19 is a small dataset consisting of 317 high-resolution ($\sim4000 \times 3000$) chest X-rays images \citep{dataset-covid, cohen2020covid, cohen2020covidProspective}. This dataset corresponds to a three-class classification problem.
\end{itemize}
Both DR and Histopathology cancer datasets are segregated into four training datasets of sizes: 500, 1000, 5000, and 10000; and a common test dataset of 2000 images. The \textit{Covid-19} dataset is partitioned into 60/20 train/validation split and a separate 20\% test set for evaluation. The images in all the datasets are resized to $224 \times 224$, which is the standard input resolution for ResNet architectures.

\subsubsection{Implementation Details}
\label{subsec:implementationdetails}
\noindent\textbf{Architectures \textendash} Due to the popularity of ResNet architectures in medical imaging for classification tasks \citep{google-medical-imaging, pretraining-medicalimaging, radimagenet}, we choose the standard ResNet18, ResNet50 \citep{resnet}, and WideResNet \citep{wideresnet} architectures as the network backbone to simulate small, medium, and large architecture sizes, respectively.
For the training regimes relying on a pretrained model, we initialize the backbone architectures using ImageNet-pretrained weights, and the classifier and self-supervised modules using the Kaiming uniform initialization \citep{kaminghe} variant.\\
\textbf{Evaluation Metrics \textendash} We use two performance metrics - \textit{Accuracy} and \textit{Area under the Receiver Operating Characteristic curve (ROC AUC)}; and six calibration metrics - \textit{ECE, MCE, ACE, OE, Brier} and \textit{NLL}.
The architecture details and hyperparameter settings are presented in the supplementary material Section \ref{ssec:sup_hyperparameters}.

\section{Results}
\label{sec:results}
\subsection{Effect of Training Regimes on Calibration}
\label{sec:effectoftrainingregimes}
\noindent In this study, we investigate the performance and calibration of three different architectures - \textit{ResNet18, ResNet50 \& WideResNet} using three different training regimes -  \textit{Fully-Supervised with random initialization} ($FS_r$), \textit{Fully-Supervised with pretraining} ($FS_p$) and \textit{Rotation-based Self-Supervision with pretraining} ($SSL_p$).

For medical image analysis, both the accuracy and reliability of the models are crucial. In this context, there are two key scenarios we need to consider:
\begin{enumerate}
    \item \textit{High accuracy and high calibration error \textendash } When a model has high accuracy but is miscalibrated, the model's predictions may not be trustworthy. Both incorrect predictions with high confidence and correct predictions with low confidence are detrimental in healthcare applications. Reliance on accuracy alone is hazardous.
    \item \textit{High accuracy and low calibration error \textendash } This is the ideal scenario, where a model has high accuracy and well-calibrated confidence scores. Predictions from such a model can be trusted in the decision-making process.
\end{enumerate}

\subsubsection{Effect of Architecture and Dataset Size}
\label{subsec:acrossdataset}
\begin{figure}[b] 
    % \centering
    \centerline{
    \includegraphics[width=\linewidth,keepaspectratio]{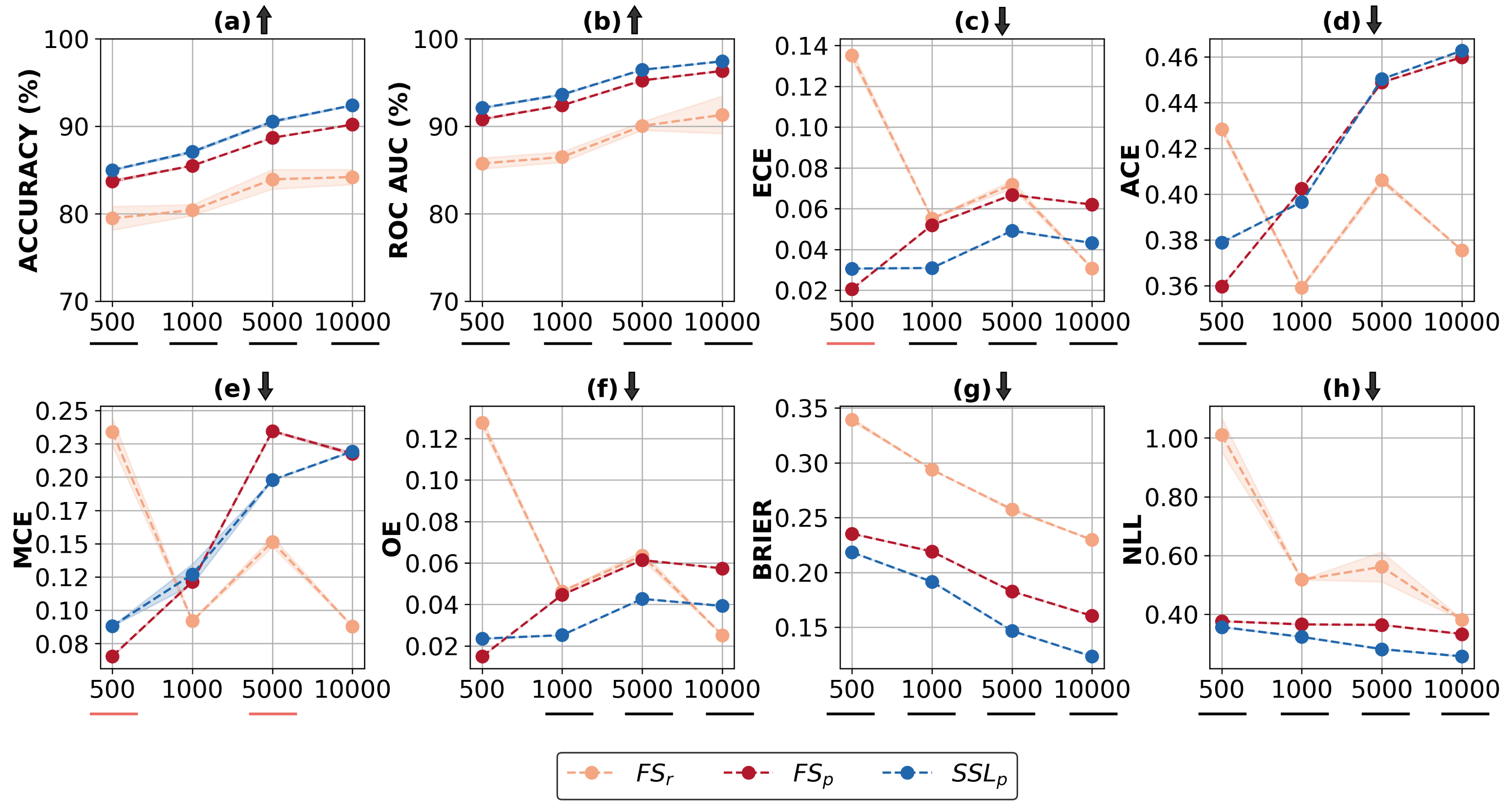}}
    \caption{Joint evaluation for performance and calibration across different dataset sizes (x-axis) using WideResNet architecture on Histopathology dataset. The shaded region corresponds to $\mu \pm \sigma$, estimated over 3 trials. $\uparrow$: higher is better, $\downarrow$: lower is better.}
    \label{fig:histopathology_wideresnet_metrics}
\end{figure}
\begin{figure*}[p]
    \centering    \includegraphics[width=.85\linewidth,keepaspectratio]{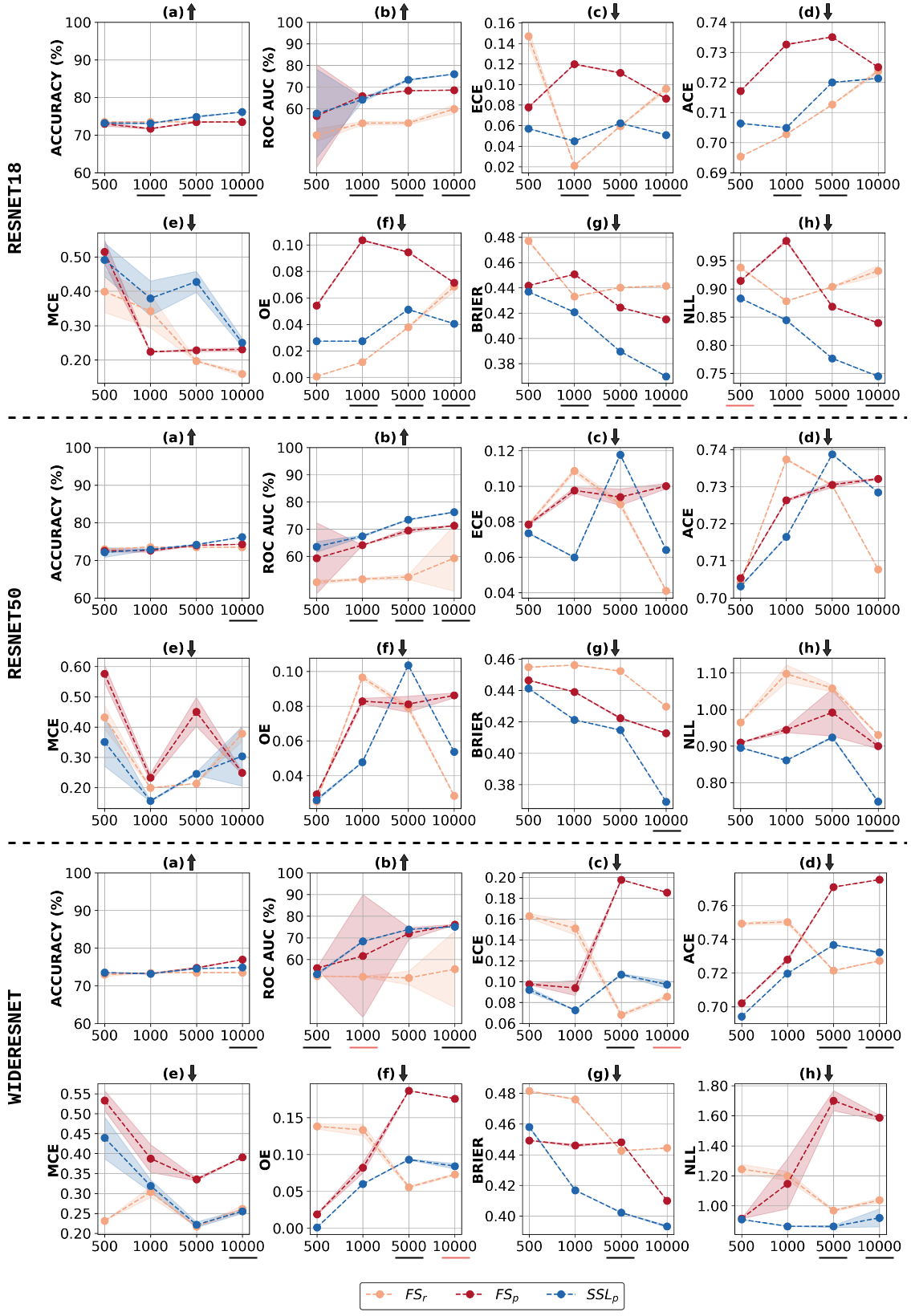}
    \caption{Joint evaluation for performance and calibration across different dataset sizes (x-axis) and architectures for DR dataset. The shaded region corresponds to $\mu \pm \sigma$, estimated over 3 trials. The underline shows the statistical significance between $FS_p$ and $SSL_p$. Black and Pink color signifies $p<0.05$ and $0.05<p<0.1$ level of significance, respectively. $\uparrow$: higher is better, $\downarrow$: lower is better.}
    \label{fig:dr_datasizes_and_arch_metrics}
\end{figure*}

\noindent In this section, we present the findings of our analysis of the DR dataset. The performance and calibration scores of various architectures, as well as the effects of increasing training dataset size, are depicted in Figure \ref{fig:dr_datasizes_and_arch_metrics} for the three different training regimes. Similar results and analysis of WideResNet architecture on the Histopathology dataset is presented in Figure \ref{fig:histopathology_wideresnet_metrics} and rest can be found in the supplementary material (Figure \ref{fig:histopathology_datasizes_and_arch}).
Owing to the difficulty of the task, the performance of all training regimes across all the models is not very high ($\leq$ 75\%). However, we do see a clear improvement in performance as the training dataset size increases across all architectures and regimes. Additionally, we observe that initializing models with pretrained weights (with $SSL_p$ having an edge over $FS_p$) offer a significant advantage over random initialization, which contradicts existing assumptions that transfer learning from ImageNet models is not beneficial. Both $FS_p$ and $SSL_p$ result in similar performance when using larger models \citep{transfusion}.

Comparing the effect of $FS_p$ and $SSL_p$ training regimes on calibration, we see that $SSL_p$ significantly improves calibration across all metrics for all architectures and training dataset sizes as illustrated in Figures \ref{fig:dr_datasizes_and_arch_metrics}(c)-(h). The gap in the calibration metrics for $SSL_p$ and $FS_p$ is highest when using the largest architecture (WideResNet). While a randomly initialized model ($FS_r$) results in marginally better calibration (sometimes even better than $SSL_p$), the performance is significantly poor. Overall, we observe that models trained using self-supervision with pretrained weights show better or similar performance with a significant improvement in calibration error compared to fully-supervised pretraining. These results suggest that self-supervised training can help improve both performance and calibration, leading to more robust and reliable models for medical image analysis.

We discuss the results on the Covid-19 dataset separately owing to its small size. Figure \ref{fig:covid_datasizes_and_arch_metrics} depicts that all the models result in high performance on this dataset indicating the ease of learning the task. The superior performance of $FS_p$ and $SSL_p$ indicate a definite advantage of transfer through pretrained over random initialization, contradicting the recent findings \citep{transfusion}. It is also evident that larger models result in better performance than shallow models. The negative impact of training from a random initialization ($FS_r$) for over-parameterized models is also evident from the drop in the performance and calibration with the increase in architecture size. While we observe a significant difference in the performance, there is only a marginal change in the calibration metrics. There is no definite trend in the calibration across the three training regimes. Thus, while transfer seems to have a positive impact on performance, calibration does not enjoy a commensurate impact.

\subsubsection{Issues with using Single Calibration Metric}
\label{subsec:single-metric}
\begin{figure*}[t]
    % \centering
    \centerline{
    \includegraphics[width=.7\linewidth,keepaspectratio]{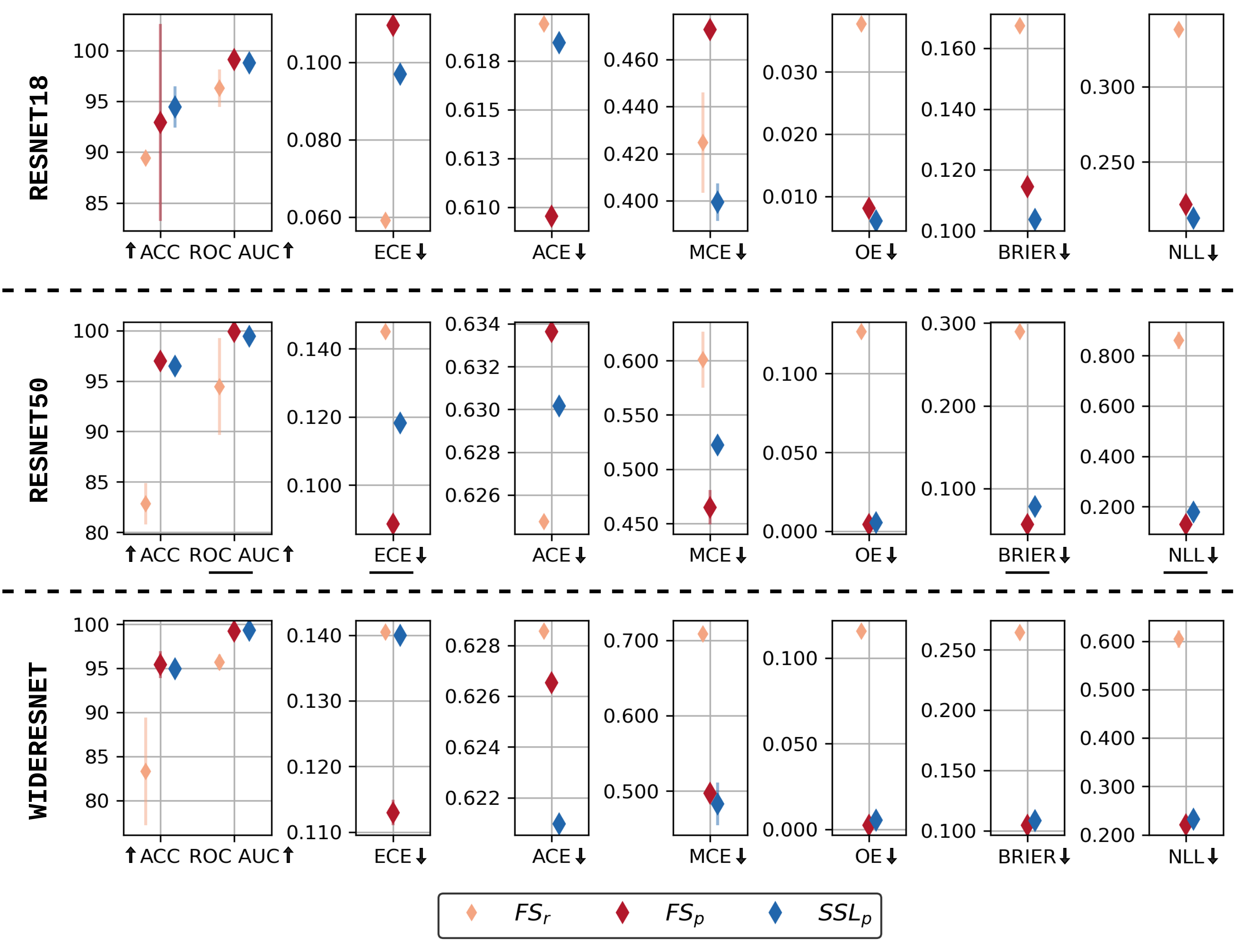}}
     \caption{Comparing performance and calibration across different architectures and training regimes for \textit{Covid-19} dataset. The error bars correspond to $\mu \pm \sigma$, estimated over 3 trials. Relying on a single calibration error metric, such as ECE or ACE, can lead to conflicting conclusions when it comes to model selection. By considering a combination of metrics, we gain a more comprehensive understanding of the model's calibration performance. $\uparrow$: higher is better, $\downarrow$: lower is better.}
    \label{fig:covid_datasizes_and_arch_metrics}
\end{figure*}
\noindent In this section, we discuss the importance of collective evaluation of calibration metrics. For this purpose, let’s consider the question - \textit{Does transfer learning improve calibration?} In the context of DR dataset, we analyze the results in Figure \ref{fig:dr_datasizes_and_arch_metrics}. Comparing $FS_r$ and $FS_p$ using only \textit{Brier} for all architectures and dataset sizes, the general trend we observe is that transfer learning improves calibration. However, this observation fails when we chose \textit{ECE} metric, which gives us mixed results. Similarly, incorrect conclusions could be drawn when using individual metrics like \textit{NLL} and \textit{ACE}.

Likewise, we consider the effect of architecture on performance and calibration in the context of the small Covid-19 dataset. From Figure \ref{fig:covid_datasizes_and_arch_metrics}, we observe that $FS_p$ and $SSL_p$ have comparable performances with nominal improvement with increasing architecture size. In this case, using only \textit{ECE} as the calibration metric would lead us to infer that $FS_p$ provides better calibration than $SSL_p$ for large capacity models. In contrast, \textit{ACE} suggests the opposite. However, these two training regimes are quite similar across most other metrics.

These examples further highlight that in scenarios where models provide mixed calibration results, selecting the best model is non-trivial/subjective. In section \ref{sec:discussion}, we discuss some potential model selection criteria to address this issue.

\subsection{Factors affecting Performance and Calibration}
\label{sec:weightdistribution}
\noindent In this section, we explore two potential factors linked to the enhanced calibration of the self-supervised training regime. Firstly, we examine the standard deviation of weight distributions and calibration metrics across different training regimes. Secondly, we investigate the similarity of learned representations in the activations.

\subsubsection{Weight Distribution}
\label{subsubsec:weightdistribution}
\noindent The weight distribution of a neural network can provide useful insights into the model's performance. Regularization schemes like $\mathcal{L}_1$, $\mathcal{L}_2$, dropout \citep{l1-l2, dropout} are often employed to find optimal parameters of a model with low generalization error. By adding a parameter norm penalty term to the objective function, the $\mathcal{L}_1$ and $\mathcal{L}_2$ norms encourage sparse weights with many zero values and small weight values respectively. Weighting the contribution of the penalty term controls the regularization effect. For instance, with $\mathcal{L}_2$ norm, the histogram of weights tends to a zero-mean normal distribution with a high penalty that causes the model to underestimate the weights and hence leads to underfitting. In contrast, a low penalty yields a flatter histogram that causes the model to overfit the training data. To strike the right balance, careful hyperparameter tuning is needed to determine the data-dependent optimal penalty term contribution for better generalization. Based on this intuition, we attempt to interpret the performance and calibration of networks trained using different regimes using weight distribution analysis. To the best of our knowledge, the calibration of a model has not been explained in the context of the weight distribution of a network, especially for medical image analysis.

\begin{figure}[t]
     \centering
     \begin{subfigure}[b]{0.49\textwidth}
         \centering
         \includegraphics[width=\linewidth,keepaspectratio]{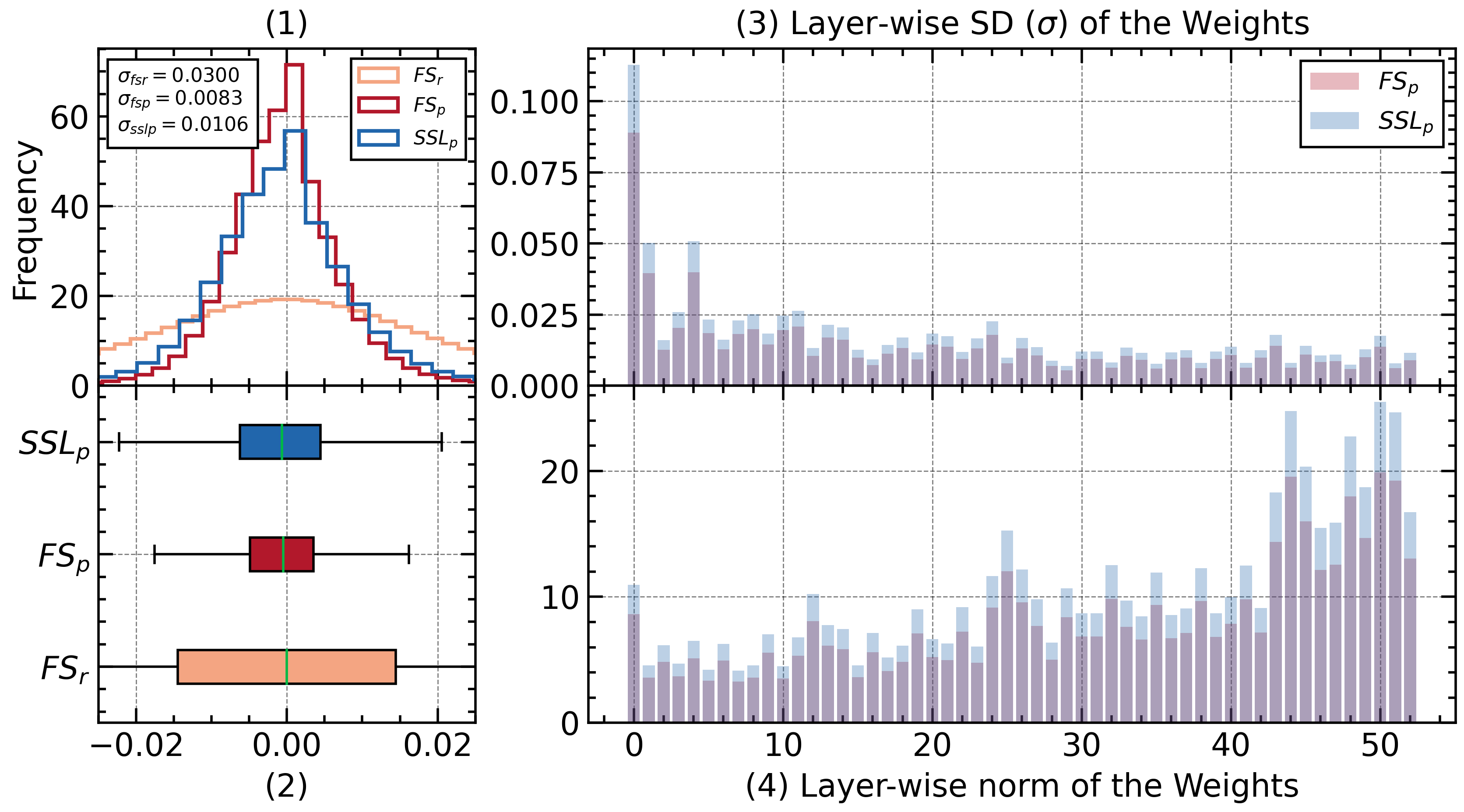}
         \caption{}
         \label{subfig:dr_weight_distribution}
     \end{subfigure}
     \hfill
     \begin{subfigure}[b]{0.49\textwidth}
         \centering
         \includegraphics[width=\linewidth,keepaspectratio]{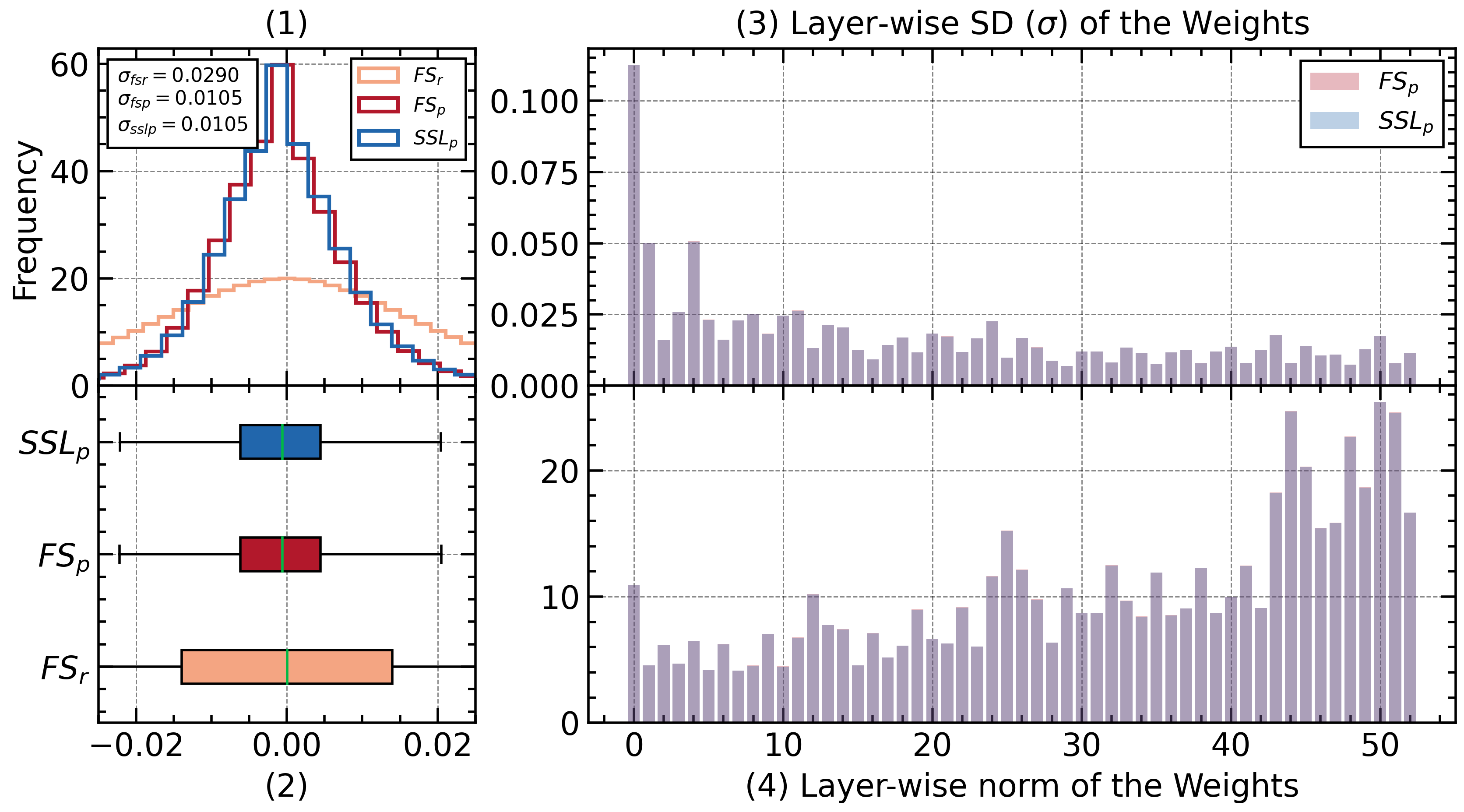}
         \caption{}
         \label{subfig:histopathology_weight_distribution}
     \end{subfigure}
        \caption{Comparing different aspects of WideResNet learned weights for dataset size 10000 on DR-(a) and Histopathology Cancer-(b) datasets. (1) and (2) the normalized histogram of weights of three training regimes. (3) Layer-wise comparison of standard deviation (SD) between $FS_p$ and $SSL_p$. (4) Layer-wise comparison of Frobenius norm between $FS_p$ and $SSL_p$.}
        \label{fig:dr_histopathology_wideresnet_weight_dist}
\end{figure}

The comparison of weight distributions between the models trained using $FS_r$, $FS_p$, and, $SSL_p$ for the DR dataset in Figure \ref{subfig:dr_weight_distribution}-(1),(2) reveals some interesting observations. The weight distribution of the model trained with $FS_p$ exhibits a higher peak than $SSL_p$, indicating that most of the weights are small. Conversely, the $FS_r$ model exhibits the highest standard deviation, resembling a uniform distribution. Now, the question arises: which distribution is preferable, and which scenario leads to better generalization with improved calibration? To address this, we analyze the impact of weight distribution on the performance and calibration of $FS_p$ and $SSL_p$ models using Figure \ref{fig:dr_datasizes_and_arch_metrics} and Figure \ref{fig:dr_histopathology_wideresnet_weight_dist}. We observe that both models show similar AUC performance, with $SSL_p$ displaying a smaller peak in the weight distribution. This difference in weight distribution influences the calibration metrics, where $SSL_p$ demonstrates significantly lower calibration error across most metrics. In other words, the predicted probabilities align more closely with the true probabilities using the $SSL_p$ model.

For Histopathology dataset, the weight distribution of the $SSL_p$ model is similar to that of the $FS_p$, as seen in Figure \ref{subfig:histopathology_weight_distribution}-(1),(2). This similarity in weight distribution could be attributed to an easier task, leading to higher test performance. However, despite the similarity in weight distribution, the $SSL_p$ model still provides better-calibrated outputs compared to the $FS_p$, but the difference in calibration error between these training regimes is now smaller. Considering the standard deviation of the weight distributions, it is suggested that a balance in the spread of weights is important for achieving good performance and calibration. It is important to note that the $FS_r$ model has the highest standard deviation and comparable calibration error, it exhibits low AUC performance, making it inconsequential among other training regimes.

\begin{figure*}[t]
     \centering
     \begin{subfigure}[b]{0.45\textwidth}
         \centering
         \includegraphics[width=\textwidth,keepaspectratio]{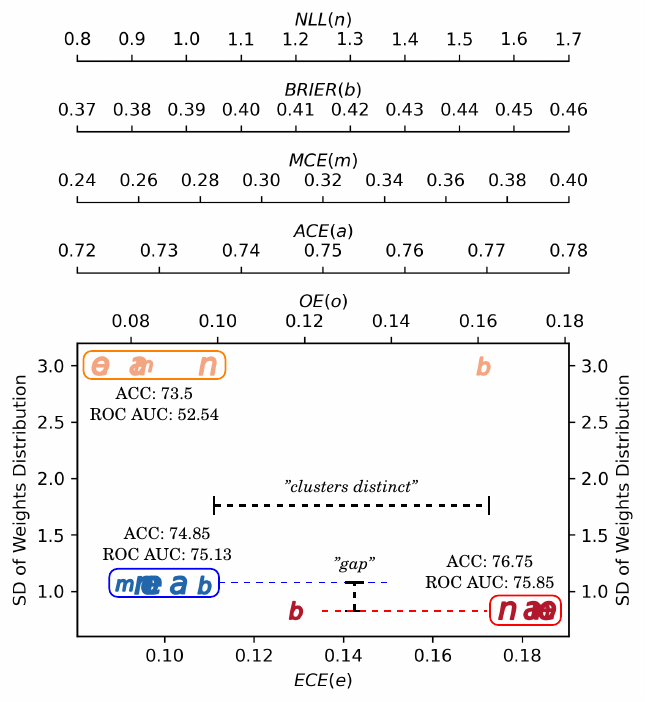}
         \caption{}
         \label{subfig:SDvsmetrics_dr}
     \end{subfigure}
     \hfill
     \begin{subfigure}[b]{0.45\textwidth}
         \centering
         \includegraphics[width=\textwidth,keepaspectratio]{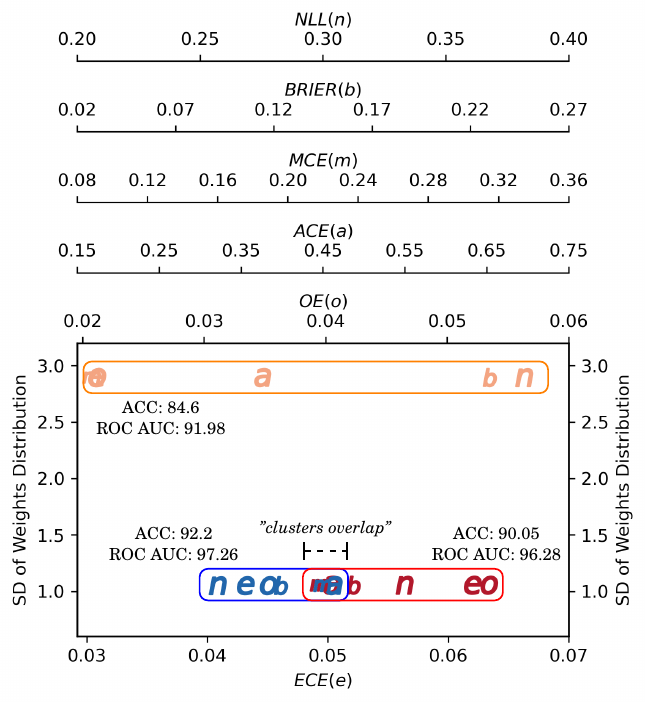}
         \caption{}
         \label{subfig:SDvsmetrics_histopathology}
     \end{subfigure}
        \caption{Comparing calibration metrics (x-axis) vs. standard deviation (SD, y-axis) of WideResNet architecture for dataset size 10000 on DR and Histopathology cancer datasets. Colors represent training regimes (orange for $FS_r$, blue for $SSL_p$, and red for $FS_p$), and markers are the lowercase initials of each calibration metric; $e$ \textendash{} \underline{E}CE, $o$ \textendash{} \underline{O}E, $a$ \textendash{} \underline{A}CE, $m$ \textendash{} \underline{M}CE, $b$ \textendash{} \underline{B}rier, $n$ \textendash{} \underline{N}LL. Alongside each calibration error cluster, the performance is also reported. Ideally, the metrics should be at the bottom left with comparable performance. \textbf{(a)} $SSL_p$ has less calibration error with on-par performance than $FS_p$ training regime, indicating it to be a suitable choice. Calibration error metrics clusters of $SSL_p$ and $FS_p$ are noticeably well separated, correlating with the gap in their SD. \textbf{(b)} Here, $SSL_p$ seems to be the best in calibration and performance compared to other training regimes. The noticeable difference we observed here is that the calibration error metrics clusters of $SSL_p$ and $FS_p$ are close (somewhat overlapping) when the SD of their weight distributions are similar.}
        \label{fig:SDvsmetrics}
\end{figure*}

In Figure \ref{fig:dr_histopathology_wideresnet_weight_dist}-(3),(4), we analyze the layer-wise standard deviation and Frobenius norm of the weights. In Figure \ref{subfig:dr_weight_distribution}, we observe $SSL_p$ influence on the standard deviation and weight magnitudes in every layer of the network. Additionally, we notice that the standard deviation tends to be higher in the initial layers and decreases as we move towards higher layers of the network. In Figure \ref{subfig:histopathology_weight_distribution}, the standard deviation and magnitude of weights are similar for both $SSL_p$ and $FS_p$ training regimes. This suggests that the features extracted by each layer of the network are similar, which could be attributed to the high performance achieved by both training regimes. Despite the similarity, the $SSL_p$ training regime still produces a better-calibrated model than the $FS_p$, indicating the additional benefits of self-supervised training.
For a more comprehensive analysis, Figure \ref{fig:SDvsmetrics} further consolidates the trends between performance, the standard deviation of the weights, and model calibration. The figure highlights that achieving good performance and calibration in a model necessitates finding a balance in the spread of weights, a balance which the $SSL_p$ training regime was able to achieve successfully. Due to the different scales of the calibration metrics, we plot them on multiple axes. The weight values and their standard deviation are very small; therefore, we scaled them by $10^2$. In Figure \ref{subfig:SDvsmetrics_dr}, $FS_r$ (top left, orange) has the highest standard deviation (wide distribution) and gives us the best calibration error (x-axis) but the worst performance compared to other training regimes. The standard deviation for $FS_p$ (bottom right, red) is the lowest, but the calibration error is still high, which is not ideal. On the other hand, $SSL_p$ has a low standard deviation but yields the best performance and calibration. So, when we encounter the  gap in the standard deviation of weights between different training regimes ($SSL_p$ and $FS_p$), we observe the calibration error metrics are well separated (Figure \ref{subfig:SDvsmetrics_dr}). Alternatively, when the gap is negligible, the calibration error metrics overlap (Figure \ref{subfig:SDvsmetrics_histopathology}). In summary, we observe that the $SSLp$ training regime consistently provides better calibration than the $FS_p$ regime for both datasets. The magnitude of improvement or change in calibration is directly related to the differences in weight distributions.

\subsubsection{Learned Representation}
\noindent In addition to the diversity of the whole weight space, we explore the impact of layer-wise, learned neural representations on performance and calibration. Towards this end, we use the widely popular Centered Kernel Alignment (CKA) \citep{cka} metric that measures the similarity between the activations of hidden layers in a neural network. Literature suggests that high representational similarity across layers indicates redundancy in learned representations of a network. Furthermore, redundant representations impact the  generalizability due to the influence of regularized training \citep{redundant-representations}, which in turn improves the model calibration \citep{ece-guo}.

\begin{figure*}[t] 
    % \centering
    \centerline{
    \includegraphics[width=.75\linewidth,keepaspectratio]{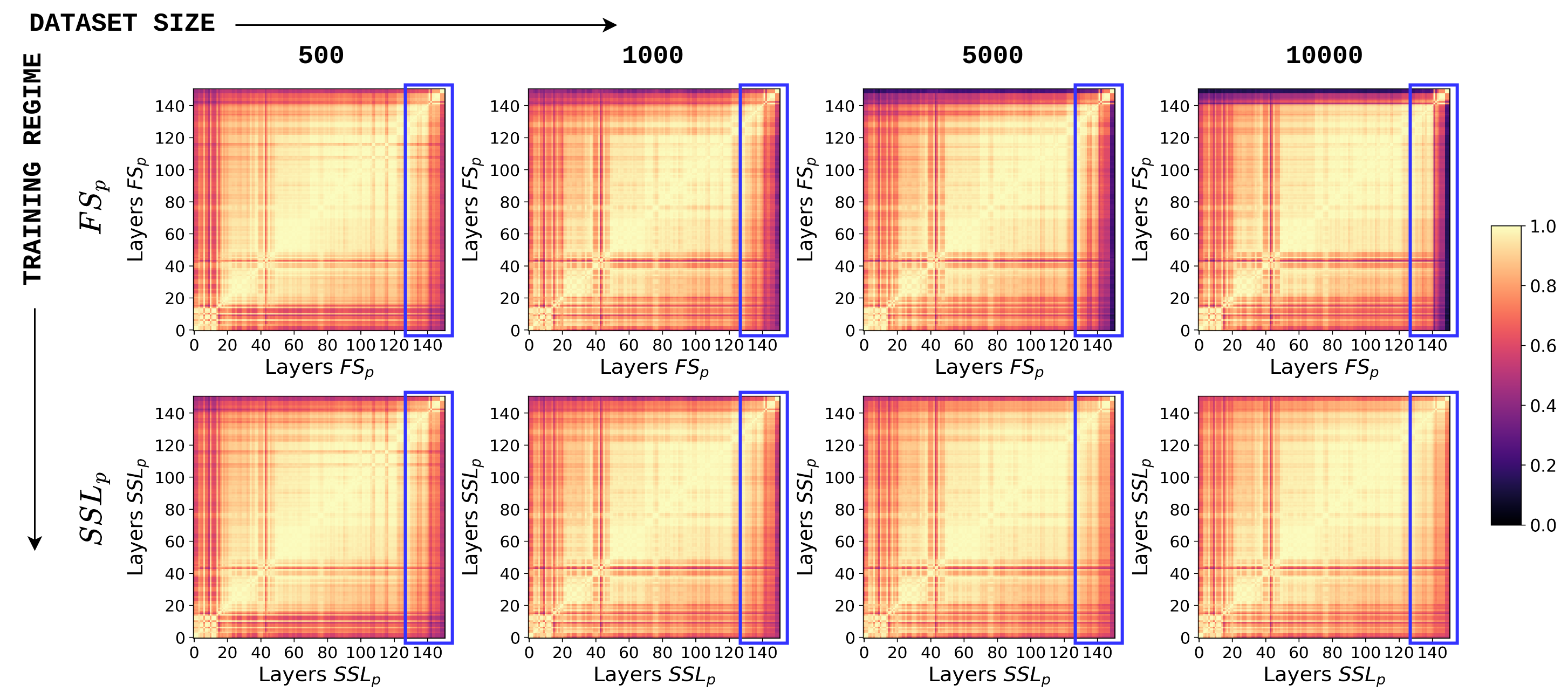}}
     \caption{CKA plots of trained WideResNet architecture using fully-supervised (pretrained, $FS_p$) and self-supervised (pretrained, $SSL_p$) regime for DR dataset. The plots represents similarity between representations of features. The range of the CKA metric is between 0 and 1, with 0 indicating two completely distinct activations (not similar) and 1 indicating two identical activations (similar).}
    \label{fig:dr_cka_wideresnet}
\end{figure*}

CKA analysis of WideResNet's layer representations for different training regimes on the DR dataset is shown in Figure \ref{fig:dr_cka_wideresnet}. The CKA plots for $FS_p$ and $SSL_p$ depict comparatively similar patterns. However, the higher layers of $FS_p$ show a significant decrease in representational similarity (darker region shown in blue box) with increasing dataset size. The relatively high CKA values of the deeper layers of $SSL_p$ depict redundancy of learned representations lighter regions) that provides implicit regularization. This in turn explains the reduced calibration error of $SSL_p$ compared to $FS_p$ as seen in Figure \ref{fig:dr_datasizes_and_arch_metrics}. A similar pattern is observed for ResNet18 and ResNet50 architectures as depicted in Figure \ref{fig:dr_cka_resnet18_resnet50} of the supplementary material. For the Histopathology dataset, the CKA plots (shown in Figure \ref{fig:histopathology_cka_all_arch} of the supplementary material) for $FS_p$ and $SSL_p$ depict very similar patterns that explain comparable performance and calibration afforded by these training regimes.

To facilitate quantitative comparison, we present the mean CKA value as a summary statistic to represent the CKA plots of individual networks in Table {\ref{tab:mean_cka}} of the Supplementary Material, Section {\ref{subsec:cka_comparision}}. While not very significant, these findings align with the trends observed in Figure {\ref{fig:dr_cka_wideresnet}}. Furthermore, the difference in the mean CKA values of $SSL_p$ and $FS_p$ fairly correlates with the difference in the magnitude of the calibration metrics of these regimes.

% \begin{figure*}[htbp]

\section{Discussion}
\label{sec:discussion}
\noindent For safety-critical applications like medical image analysis, it is imperative to choose models with high accuracy and low calibration errors. In this study, we investigate the performance and calibration of three different architectures using three different training regimes on medical imaging datasets of varying sizes and task complexities. Furthermore, we use six complementary calibration metrics that collectively provide a comprehensive evaluation of the predictive uncertainty of the models.

\textit{Model selection with mixed calibration results \textendash} While using multiple calibration metrics provides a more comprehensive evaluation, deciding on the best model can still be challenging as observed in Section \ref{subsec:single-metric}. There are a few strategies that can be employed to aid in the decision-making process. One approach is to use a voting-based scheme, where each model is assigned a vote based on its performance across the calibration metrics. The model with the maximum number of votes is then selected as the best choice. This approach treats all metrics equally and can be useful when there is no significant variation in the importance of different metrics.

\textit{Domain specific metric relevance \textendash} However, it is important to consider that different calibration metrics may have different objectives and importance in specific domains. For example, metrics like OE (Overconfidence Error) explicitly measure the overconfidence of the model predictions, while MCE (Maximum Calibration Error) provides an upper bound on the mistakes made by the model. In such cases, it might be necessary to assign more weightage to these important metrics during the voting process. The determination of metric importance is subjective and can vary depending on the application. Expert knowledge and domain expertise play a crucial role in assigning relative importance to different metrics. By incorporating the opinions of experts, the voting process can be tailored to reflect the specific requirements of the application.

\textit{Margin for model selection \textendash} In addition to assigning weights to metrics, introducing a margin or threshold in the voting scheme can help refine the model selection process. This threshold represents the minimum difference in calibration error between two training regimes that must be surpassed for a metric to be considered in the model selection. By setting a threshold, the metrics can be filtered out that do not exhibit significant differences and focus on those that have a substantial impact on model calibration.

It is worth noting that the difficulty of choosing a model also arises when one model has higher accuracy but poorer calibration while another model has lower accuracy but better calibration. This dilemma has been discussed in the literature \citep{calibration-revisiting}, highlighting the need for careful consideration of calibration metrics during model selection.
\textit{Selective prediction} is one scenario where we abstain the classifier that gives us low-confident predictions based on some threshold or cost structure of the specific application \citep{selective-prediction-thresh1}. In such cases, low-confidence predictions are referred to an expert for further analysis or diagnosis. This approach allows for cautious decision-making when the model's confidence is not sufficient for reliable predictions. Overall, the selection of the best model with mixed calibration results requires a combination of objective evaluation, subjective judgment of metric importance, and consideration of domain-specific requirements.

\textit{Calibration Metrics \textendash}
While we have elaborated on the drawbacks of ECE, it provides an intuitive and straightforward interpretation, is simple to implement, and captures pure calibration. Additionally, ECE is associated with the reliability diagram - a powerful tool to visualize model calibration. It's also worth noting that alternative calibration metrics have their own shortcomings. The majority of the existing metrics suffer from challenges like scale-dependent interpretation, lack of normalized range, arbitrary choice of number of bins, etc. \citep{tce}. Moreover, composite measures like NLL and Brier blend calibration and refinement, making it challenging to isolate calibration effects. Multiclass settings introduce additional complexity due to the multitude of classes, their diverse interrelations, and the absence of a universally accepted metric for gauging refinement. Moreover, the choice of calibration metric can also be domain or application-dependent. As there is no universally applicable or acceptable calibration metric, we proposed collective evaluation of these metrics for a better or unbiased understanding of calibration performance.

\textit{Limitations \textendash} Our current study focused on medical image classification tasks across three different benchmark datasets. However, due to limited computational resources, we selected datasets with 2D images. Extending this work to 3D datasets as well as other tasks like medical image segmentation and registration, can help broaden our understanding of calibration in the general context of medical image analysis. Additionally, our study highlights that using the rotation-based self-supervised learning (SSL) approach gives better-calibrated results compared to the usual fully-supervised learning. A comparison of other SSL techniques, such as contrastive SSL or generative SSL, would be interesting.

\textit{Conclusion \textendash} In general, for medical image classification tasks, we observe that training regimes have a varying impact on model calibration. Overall, we observe that across different architectures, training regimes, datasets, and sample sizes, (a) transfer learning through pretraining helps improve performance over random-initialized models and (b) pretrained self-supervised approach provides better calibration than its fully supervised counterpart, with on-par or better performance. While we notice a sizeable increase in performance with dataset sizes, only nominal improvement is realized with increasing model capacity.

Furthermore, we identified weight distribution and learned representations of a neural network as potential confounding factors that provide useful insights into model calibration, in particular, to explain the superiority of a rotation-based self-supervised training regime over fully supervised training. 

\textit{Broader Impact \textendash} We anticipate that this analysis will offer significant insights into calibration across datasets of varying sizes and models of different complexities. This work raises a broader question regarding the search for a unified metric that can provide a comprehensive understanding of model calibration, thereby reducing the need to evaluate models based on multiple criteria. Ensuring accurate and reliable probabilistic predictions is vital for effective risk management and decision-making. It is particularly important when relying on the outputs of probabilistic models that require trust. Additionally, developing well-calibrated models is essential for promoting the widespread acceptance of machine learning methods, especially in fields like AI-driven medical diagnosis, as it directly influences the level of trust in new technologies and improves their explainability.

\section*{Acknowledgment}
\noindent The support and the resources provided by PARAM Sanganak under the National Supercomputing Mission, Government of India at the Indian Institute of Technology, Kanpur are gratefully acknowledged.

%% The Appendices part is started with the command \appendix;
%% appendix sections are then done as normal sections
%% \appendix

%% \section{}
%% \label{}

%% If you have bibdatabase file and want bibtex to generate the
%% bibitems, please use
%%
% \bibliographystyle{elsarticle-harv} 
\bibliography{reference}

\begin{thebibliography}{78}
\providecommand{\natexlab}[1]{#1}
\providecommand{\url}[1]{\texttt{#1}}
\expandafter\ifx\csname urlstyle\endcsname\relax
  \providecommand{\doi}[1]{doi: #1}\else
  \providecommand{\doi}{doi: \begingroup \urlstyle{rm}\Url}\fi

\bibitem[Azizi et~al.(2021)]{google-medical-imaging}
Shekoofeh Azizi et~al.
\newblock Big self-supervised models advance medical image classification.
\newblock In \emph{IEEE International Conference on Computer Vision (ICCV)},
  2021.

\bibitem[Blundell et~al.(2015)Blundell, Cornebise, Kavukcuoglu, and
  Wierstra]{bnn}
Charles Blundell, Julien Cornebise, Koray Kavukcuoglu, and Daan Wierstra.
\newblock Weight uncertainty in neural networks.
\newblock In \emph{International Conference on Machine Learning (ICML)}, 2015.

\bibitem[Brier(1950)]{calibration-brier}
Glenn~W. Brier.
\newblock Verification of forecasts expressed in terms of probability.
\newblock \emph{Monthly Weather Review}, 1950.

\bibitem[Caruana et~al.(2015)Caruana, Lou, Gehrke, Koch, Sturm, and
  Elhadad]{caruana}
Rich Caruana, Yin Lou, Johannes Gehrke, Paul Koch, Marc Sturm, and Noemie
  Elhadad.
\newblock Intelligible models for healthcare: Predicting pneumonia risk and
  hospital 30-day readmission.
\newblock In \emph{ACM SIGKDD International Conference on Knowledge Discovery
  and Data Mining}, 2015.

\bibitem[Cohen et~al.(2020{\natexlab{a}})Cohen, Morrison, and
  Dao]{cohen2020covid}
Joseph~Paul Cohen, Paul Morrison, and Lan Dao.
\newblock Covid-19 image data collection.
\newblock \emph{arXiv 2003.11597}, 2020{\natexlab{a}}.
\newblock URL \url{https://github.com/ieee8023/covid-chestxray-dataset}.

\bibitem[Cohen et~al.(2020{\natexlab{b}})Cohen, Morrison, Dao, Roth, Duong, and
  Ghassemi]{cohen2020covidProspective}
Joseph~Paul Cohen, Paul Morrison, Lan Dao, Karsten Roth, Tim~Q Duong, and
  Marzyeh Ghassemi.
\newblock Covid-19 image data collection: Prospective predictions are the
  future.
\newblock \emph{arXiv 2006.11988}, 2020{\natexlab{b}}.
\newblock URL \url{https://github.com/ieee8023/covid-chestxray-dataset}.

\bibitem[{Covid-19 Image Dataset}()]{dataset-covid}
{Covid-19 Image Dataset}.
\newblock
  \url{https://www.kaggle.com/datasets/pranavraikokte/covid19-image-dataset}.

\bibitem[Doersch et~al.(2015)Doersch, Gupta, and Efros]{ssl-github}
Carl Doersch, Abhinav Gupta, and Alexei~A. Efros.
\newblock Unsupervised visual representation learning by context prediction.
\newblock In \emph{IEEE International Conference on Computer Vision (ICCV)},
  2015.

\bibitem[Doimo et~al.(2022)Doimo, Glielmo, Goldt, and
  Laio]{redundant-representations}
Diego Doimo, Aldo Glielmo, Sebastian Goldt, and Alessandro Laio.
\newblock Redundant representations help generalization in wide neural
  networks.
\newblock In \emph{Neural Information Processing Systems (NeurIPS)}, 2022.

\bibitem[Donahue et~al.(2014)Donahue, Jia, et~al.]{pretraining-decaf}
Jeff Donahue, Yangqing Jia, et~al.
\newblock Decaf: A deep convolutional activation feature for generic visual
  recognition.
\newblock In \emph{International Conference on Machine Learning (ICML)}, 2014.

\bibitem[Ehteshami~Bejnordi et~al.(2017)]{dataset-histopathology2}
Babak Ehteshami~Bejnordi et~al.
\newblock {Diagnostic Assessment of Deep Learning Algorithms for Detection of
  Lymph Node Metastases in Women With Breast Cancer}.
\newblock \emph{JAMA}, 2017.

\bibitem[Ericsson et~al.(2021)Ericsson, Gouk, and Hospedales]{ssl_transfer}
Linus Ericsson, Henry Gouk, and Timothy~M. Hospedales.
\newblock How well do self-supervised models transfer?
\newblock In \emph{IEEE Conference on Computer Vision and Pattern Recognition
  (CVPR)}, 2021.

\bibitem[{EyePACS, Diabetic Retinopathy Detection}()]{dataset-eyepacs}
{EyePACS, Diabetic Retinopathy Detection}.
\newblock
  \url{https://www.kaggle.com/competitions/diabetic-retinopathy-detection/}.

\bibitem[Fong and Vedaldi(2017)]{xai-mp}
Ruth~C. Fong and Andrea Vedaldi.
\newblock Interpretable explanations of black boxes by meaningful perturbation.
\newblock \emph{IEEE International Conference on Computer Vision (ICCV)}, 2017.

\bibitem[Frenkel and Goldberger(2022)]{weightscaling-miccai}
Lior Frenkel and Jacob Goldberger.
\newblock Calibration of medical imaging classification systems with weight
  scaling.
\newblock In \emph{Medical Image Computing and Computer Assisted Intervention
  (MICCAI)}, 2022.

\bibitem[Gal and Ghahramani(2016)]{mc-dropout}
Yarin Gal and Zoubin Ghahramani.
\newblock Dropout as a bayesian approximation: Representing model uncertainty
  in deep learning.
\newblock In \emph{International Conference on Machine Learning (ICML)}, 2016.

\bibitem[Gidaris et~al.(2018)Gidaris, Singh, and Komodakis]{rotationnet}
Spyros Gidaris, Praveer Singh, and Nikos Komodakis.
\newblock Unsupervised representation learning by predicting image rotations.
\newblock In \emph{International Conference on Learning Representations
  (ICLR)}, 2018.

\bibitem[Girshick et~al.(2014)Girshick, Donahue, Darrell, and
  Malik]{pretraining-rcnn}
Ross Girshick, Jeff Donahue, Trevor Darrell, and Jitendra Malik.
\newblock Rich feature hierarchies for accurate object detection and semantic
  segmentation.
\newblock In \emph{IEEE Conference on Computer Vision and Pattern Recognition
  (CVPR)}, 2014.

\bibitem[Gneiting and Raftery(2007)]{brier-proper}
Tilmann Gneiting and Adrian~E Raftery.
\newblock Strictly proper scoring rules, prediction, and estimation.
\newblock \emph{Journal of the American Statistical Association (JASA)}, 2007.

\bibitem[Guo et~al.(2017)Guo, Pleiss, Sun, and Weinberger]{ece-guo}
Chuan Guo, Geoff Pleiss, Yu~Sun, and Kilian~Q. Weinberger.
\newblock On calibration of modern neural networks.
\newblock In \emph{International Conference on Machine Learning (ICML)}, 2017.

\bibitem[He et~al.(2015)He, Zhang, Ren, and Sun]{kaminghe}
Kaiming He, X.~Zhang, Shaoqing Ren, and Jian Sun.
\newblock Delving deep into rectifiers: Surpassing human-level performance on
  imagenet classification.
\newblock \emph{IEEE International Conference on Computer Vision (ICCV)}, 2015.

\bibitem[He et~al.(2016)He, Zhang, Ren, and Sun]{resnet}
Kaiming He, Xiangyu Zhang, Shaoqing Ren, and Jian Sun.
\newblock Deep residual learning for image recognition.
\newblock In \emph{IEEE Conference on Computer Vision and Pattern Recognition
  (CVPR)}, 2016.

\bibitem[He et~al.(2017)He, Gkioxari, Dollár, and Girshick]{mask-cnn}
Kaiming He, Georgia Gkioxari, Piotr Dollár, and Ross Girshick.
\newblock Mask r-cnn.
\newblock In \emph{IEEE International Conference on Computer Vision (ICCV)},
  2017.

\bibitem[Hendrycks et~al.(2019{\natexlab{a}})Hendrycks, Lee, and
  Mazeika]{rmsce-hendrycks}
Dan Hendrycks, Kimin Lee, and Mantas Mazeika.
\newblock Using pre-training can improve model robustness and uncertainty.
\newblock In \emph{International Conference on Machine Learning (ICML)},
  2019{\natexlab{a}}.

\bibitem[Hendrycks et~al.(2019{\natexlab{b}})Hendrycks, Mazeika, and
  Dietterich]{rmsce-hendrycks2019oe}
Dan Hendrycks, Mantas Mazeika, and Thomas Dietterich.
\newblock Deep anomaly detection with outlier exposure.
\newblock In \emph{International Conference on Learning Representations
  (ICLR)}, 2019{\natexlab{b}}.

\bibitem[Hendrycks et~al.(2019{\natexlab{c}})Hendrycks, Mazeika, Kadavath, and
  Song]{ssl-hendrycks}
Dan Hendrycks, Mantas Mazeika, Saurav Kadavath, and Dawn Song.
\newblock Using self-supervised learning can improve model robustness and
  uncertainty.
\newblock \emph{Neural Information Processing Systems (NeurIPS)},
  2019{\natexlab{c}}.

\bibitem[Hendrycks et~al.(2020)Hendrycks, Mu, Cubuk, Zoph, Gilmer, and
  Lakshminarayanan]{calibration-augmix-hendrycks}
Dan Hendrycks, Norman Mu, Ekin~D. Cubuk, Barret Zoph, Justin Gilmer, and Balaji
  Lakshminarayanan.
\newblock {AugMix}: A simple data processing method to improve robustness and
  uncertainty.
\newblock \emph{International Conference on Learning Representations (ICLR)},
  2020.

\bibitem[Hern{{\'a}}ndez-Orallo et~al.(2012)Hern{{\'a}}ndez-Orallo, Flach, and
  Ferri]{selective-prediction-thresh1}
Jos{{\'e}} Hern{{\'a}}ndez-Orallo, Peter Flach, and C{{\`e}}sar Ferri.
\newblock A unified view of performance metrics: Translating threshold choice
  into expected classification loss.
\newblock \emph{Journal of Machine Learning Research}, 2012.

\bibitem[{Histopathologic Cancer Detection: Modified version of the
  PatchCamelyon (PCam) Benchmark Dataset}()]{dataset-histopathology1}
{Histopathologic Cancer Detection: Modified version of the PatchCamelyon (PCam)
  Benchmark Dataset}.
\newblock
  \url{https://www.kaggle.com/competitions/histopathologic-cancer-detection/}.

\bibitem[Jena and Awate(2019)]{bnn-awate}
Rohit Jena and Suyash~P. Awate.
\newblock A bayesian neural net to segment images with uncertainty estimates
  and good calibration.
\newblock In \emph{Information Processing in Medical Imaging (IPMI)}, 2019.

\bibitem[Jiang et~al.(2012)Jiang, Osl, Kim, and Ohno-Machado]{clinical-jiang}
Xiaoqian Jiang, Melanie Osl, Jihoon Kim, and Lucila Ohno-Machado.
\newblock Calibrating predictive model estimates to support personalized
  medicine.
\newblock \emph{Journal of the American Medical Informatics Association :
  JAMIA}, 2012.

\bibitem[Jungo and Reyes(2019)]{uncertainty-reliability-reyes}
Alain Jungo and Mauricio Reyes.
\newblock Assessing reliability and challenges of uncertainty estimations for
  medical image segmentation.
\newblock In \emph{Medical Image Computing and Computer Assisted Intervention
  -- MICCAI 2019}, 2019.

\bibitem[Jungo et~al.(2020)Jungo, Balsiger, and
  Reyes]{uncertainty-subject-dataset-level-reyes}
Alain Jungo, Fabian Balsiger, and Mauricio Reyes.
\newblock Analyzing the quality and challenges of uncertainty estimations for
  brain tumor segmentation.
\newblock \emph{Frontiers in Neuroscience}, 14, 2020.

\bibitem[Kompa et~al.(2021)Kompa, Snoek, and Beam]{calibration-second-opinion}
Benjamin Kompa, Jasper Snoek, and Andrew~L. Beam.
\newblock Second opinion needed: communicating uncertainty in medical machine
  learning.
\newblock \emph{npj Digital Medicine}, 2021.

\bibitem[Kornblith et~al.(2019)Kornblith, Norouzi, Lee, and Hinton]{cka}
Simon Kornblith, Mohammad Norouzi, Honglak Lee, and Geoffrey Hinton.
\newblock Similarity of neural network representations revisited.
\newblock In \emph{International Conference on Machine Learning (ICML)}, 2019.

\bibitem[Krizhevsky et~al.(2012)Krizhevsky, Sutskever, and
  Hinton]{pretraining-krizhevsky}
Alex Krizhevsky, Ilya Sutskever, and Geoffrey~E Hinton.
\newblock Imagenet classification with deep convolutional neural networks.
\newblock In \emph{Neural Information Processing Systems (NeurIPS)}, 2012.

\bibitem[Kruppa et~al.(2014)Kruppa, Liu, et~al.]{calibration-brier-multi}
Jochen Kruppa, Yufeng Liu, et~al.
\newblock Probability estimation with machine learning methods for dichotomous
  and multicategory outcome: Theory.
\newblock \emph{Biometrical Journal}, 2014.

\bibitem[Kull and Flach(2015)]{calibration-kull-brier-nll}
Meelis Kull and Peter Flach.
\newblock Novel decompositions of proper scoring rules for classification:
  Score adjustment as precursor to calibration.
\newblock In \emph{Machine Learning and Knowledge Discovery in Databases (ECML
  PKDD)}, 2015.

\bibitem[Lakshminarayanan et~al.(2017)Lakshminarayanan, Pritzel, and
  Blundell]{calibration-lakshminarayanan}
Balaji Lakshminarayanan, Alexander Pritzel, and Charles Blundell.
\newblock Simple and scalable predictive uncertainty estimation using deep
  ensembles.
\newblock In \emph{Neural Information Processing Systems (NeurIPS)}, 2017.

\bibitem[Langlotz et~al.(2019)]{roadmap-ai-medical-imaging}
Curtis~P. Langlotz et~al.
\newblock A roadmap for foundational research on artificial intelligence in
  medical imaging: From the 2018 nih/rsna/acr/the academy workshop.
\newblock \emph{Radiology}, 2019.

\bibitem[Larrazabal et~al.(2021)Larrazabal, Martinez, Dolz, and
  Ferrante]{meep-miccai}
Agostina~J. Larrazabal, Cesar Martinez, Jos{\'e} Dolz, and Enzo Ferrante.
\newblock Maximum entropy on erroneous predictions (meep): Improving model
  calibration for medical image segmentation.
\newblock \emph{ArXiv}, abs/2112.12218, 2021.

\bibitem[Ma et~al.(2022)Ma, He, Xi, Ebadi, Tremblay, and
  Wong]{trustworthy-scaricity-medicalimaging}
Kai Ma, Siyuan He, Pengcheng Xi, Ashkan Ebadi, St{\'e}phane Tremblay, and
  Alexander Wong.
\newblock A trustworthy framework for medical image analysis with deep
  learning.
\newblock \emph{arXiv preprint arXiv:2212.02764}, 2022.

\bibitem[Matsubara et~al.(2023)Matsubara, Tax, Mudd, and Guy]{tce}
Takuo Matsubara, Niek Tax, Richard Mudd, and Ido Guy.
\newblock {TCE}: A test-based approach to measuring calibration error.
\newblock In \emph{Uncertainty in Artificial Intelligence}, 2023.

\bibitem[Mehrtash et~al.(2020)Mehrtash, Wells, Tempany, Abolmaesumi, and
  Kapur]{uncertainty-segmentation-mehrtash}
Alireza Mehrtash, William~M. Wells, Clare~M. Tempany, Purang Abolmaesumi, and
  Tina Kapur.
\newblock Confidence calibration and predictive uncertainty estimation for deep
  medical image segmentation.
\newblock \emph{IEEE Transactions on Medical Imaging}, 2020.

\bibitem[Mei et~al.(2022)]{radimagenet}
Xueyan Mei et~al.
\newblock Radimagenet: An open radiologic deep learning research dataset for
  effective transfer learning.
\newblock \emph{Radiology: Artificial Intelligence}, 2022.

\bibitem[Minderer et~al.(2021)]{calibration-revisiting}
Matthias Minderer et~al.
\newblock Revisiting the calibration of modern neural networks.
\newblock In \emph{Neural Information Processing Systems (NeurIPS)}, 2021.

\bibitem[Mukhoti et~al.(2020)Mukhoti, Kulharia, Sanyal, Golodetz, Torr, and
  Dokania]{focalloss}
Jishnu Mukhoti, Viveka Kulharia, Amartya Sanyal, Stuart Golodetz, Philip Torr,
  and Puneet Dokania.
\newblock Calibrating deep neural networks using focal loss.
\newblock In \emph{Neural Information Processing Systems (NeurIPS)}, 2020.

\bibitem[Murphy and Winkler(1977)]{confidence-reliability}
Allan~H. Murphy and Robert~L. Winkler.
\newblock Reliability of subjective probability forecasts of precipitation and
  temperature.
\newblock \emph{Journal of the Royal Statistical Society. Series C (Applied
  Statistics)}, 1977.

\bibitem[Murugesan et~al.(2023)Murugesan, Liu, Galdran, Ayed, and
  Dolz]{devil-margin-media}
Balamurali Murugesan, Bingyuan Liu, Adrian Galdran, Ismail~Ben Ayed, and Jose
  Dolz.
\newblock Calibrating segmentation networks with margin-based label smoothing.
\newblock \emph{Medical Image Analysis}, 2023.

\bibitem[Naeini et~al.(2015)Naeini, Cooper, and Hauskrecht]{ece-naeini}
Mahdi~Pakdaman Naeini, Gregory~F. Cooper, and Milos Hauskrecht.
\newblock Obtaining well calibrated probabilities using bayesian binning.
\newblock In \emph{AAAI Conference on Artificial Intelligence}, 2015.

\bibitem[Navarro et~al.(2021)Navarro, Watanabe, et~al.]{ssl_robustness}
Fernando Navarro, Christopher Watanabe, et~al.
\newblock Evaluating the robustness of self-supervised learning in medical
  imaging.
\newblock \emph{ArXiv}, 2021.

\bibitem[Ng(2004)]{l1-l2}
Andrew~Y. Ng.
\newblock Feature selection, l1 vs. l2 regularization, and rotational
  invariance.
\newblock In \emph{International Conference on Machine Learning (ICML)}, 2004.

\bibitem[Nguyen and O{'}Connor(2015)]{emnlp_calibration}
Khanh Nguyen and Brendan O{'}Connor.
\newblock Posterior calibration and exploratory analysis for natural language
  processing models.
\newblock In \emph{Empirical Methods in Natural Language Processing (EMNLP)},
  2015.

\bibitem[Nixon et~al.(2019)Nixon, Dusenberry, Zhang, Jerfel, and
  Tran]{calibration-metrics}
Jeremy Nixon, Michael~W. Dusenberry, Linchuan Zhang, Ghassen Jerfel, and Dustin
  Tran.
\newblock Measuring calibration in deep learning.
\newblock In \emph{IEEE Conference on Computer Vision and Pattern Recognition
  (CVPR) Workshops}, 2019.

\bibitem[Petsiuk et~al.(2018)Petsiuk, Das, and Saenko]{xai-rise}
Vitali Petsiuk, Abir Das, and Kate Saenko.
\newblock Rise: Randomized input sampling for explanation of black-box models.
\newblock In \emph{British Machine Vision Conference (BMVC)}, 2018.

\bibitem[Platt(1999)]{calibration-platt}
John Platt.
\newblock Probabilistic outputs for support vector machines and comparisons to
  regularized likelihood methods.
\newblock \emph{Advances in large margin classifiers}, 1999.

\bibitem[Qui{\~{n}}onero-Candela et~al.(2006)Qui{\~{n}}onero-Candela,
  Rasmussen, Sinz, Bousquet, and Sch{\"o}lkopf]{calibration-nll}
Joaquin Qui{\~{n}}onero-Candela, Carl~Edward Rasmussen, Fabian Sinz, Olivier
  Bousquet, and Bernhard Sch{\"o}lkopf.
\newblock Evaluating predictive uncertainty challenge.
\newblock In \emph{Machine Learning Challenges. Evaluating Predictive
  Uncertainty, Visual Object Classification, and Recognising Tectual
  Entailment}, 2006.

\bibitem[Raghu et~al.(2019)Raghu, Zhang, Kleinberg, and Bengio]{transfusion}
Maithra Raghu, Chiyuan Zhang, Jon Kleinberg, and Samy Bengio.
\newblock Transfusion: Understanding transfer learning for medical imaging.
\newblock In \emph{Neural Information Processing Systems (NeurIPS)}, 2019.

\bibitem[Rahaman and thiery(2021)]{uncertainty_ensembles}
Rahul Rahaman and alexandre thiery.
\newblock Uncertainty quantification and deep ensembles.
\newblock In M.~Ranzato, A.~Beygelzimer, Y.~Dauphin, P.S. Liang, and J.~Wortman
  Vaughan, editors, \emph{Neural Information Processing Systems (NeurIPS},
  pages 20063--20075, 2021.

\bibitem[Scafarto et~al.(2023)Scafarto, Posocco, and
  Bonnefoy]{calibrate-intrepret}
Gregory Scafarto, Nicolas Posocco, and Antoine Bonnefoy.
\newblock Calibrate to interpret.
\newblock In Massih-Reza Amini, St{\'e}phane Canu, Asja Fischer, Tias Guns,
  Petra Kralj~Novak, and Grigorios Tsoumakas, editors, \emph{European
  Conference on Machine Learning and Principles and Practice of Knowledge
  Discovery in Databases (ECML PKDD)}, 2023.

\bibitem[Shrikumar et~al.(2017)Shrikumar, Greenside, and Kundaje]{xai-deeplift}
Avanti Shrikumar, Peyton Greenside, and Anshul Kundaje.
\newblock Learning important features through propagating activation
  differences.
\newblock In \emph{International Conference on Machine Learning (ICML)}, 2017.

\bibitem[Singh et~al.(2021)Singh, Bay, Sengupta, and
  Mirabile]{calibration-dark}
Aditya Singh, Alessandro Bay, Biswa Sengupta, and Andrea Mirabile.
\newblock On the dark side of calibration for modern neural networks.
\newblock In \emph{Workshop on Uncertainty and Robustness in Deep Learning
  (UDL)}, 2021.

\bibitem[Singh~Sambyal et~al.(2022)Singh~Sambyal, Krishnan, and
  Bathula]{sambyal-aleatoric-isbi}
Abhishek Singh~Sambyal, Narayanan~C Krishnan, and Deepti~R Bathula.
\newblock Towards reducing aleatoric uncertainty for medical imaging tasks.
\newblock In \emph{IEEE 19th International Symposium on Biomedical Imaging
  (ISBI)}, 2022.

\bibitem[Srivastava et~al.(2014)Srivastava, Hinton, Krizhevsky, Sutskever, and
  Salakhutdinov]{dropout}
Nitish Srivastava, Geoffrey Hinton, Alex Krizhevsky, Ilya Sutskever, and Ruslan
  Salakhutdinov.
\newblock Dropout: A simple way to prevent neural networks from overfitting.
\newblock \emph{Journal of Machine Learning Research}, 2014.

\bibitem[Stolte et~al.(2022)Stolte, Volle, Indahlastari, Albizu, Woods, Brink,
  Hale, and Fang]{domino-miccai}
Skylar~E. Stolte, Kyle Volle, Aprinda Indahlastari, Alejandro Albizu, Adam~J.
  Woods, Kevin Brink, Matthew Hale, and Ruogu Fang.
\newblock Domino: Domain-aware model calibration in medical image segmentation.
\newblock In \emph{Medical Image Computing and Computer Assisted Intervention
  (MICCAI)}, 2022.

\bibitem[Tendle and Hasan(2021)]{sslgeneralizable2021}
Atharva Tendle and Mohammad~Rashedul Hasan.
\newblock A study of the generalizability of self-supervised representations.
\newblock \emph{Machine Learning with Applications}, 2021.

\bibitem[Thulasidasan et~al.(2019)Thulasidasan, Chennupati, Bilmes,
  Bhattacharya, and Michalak]{overconfidenceerror-mixup}
Sunil Thulasidasan, Gopinath Chennupati, Jeff~A Bilmes, Tanmoy Bhattacharya,
  and Sarah Michalak.
\newblock On mixup training: Improved calibration and predictive uncertainty
  for deep neural networks.
\newblock In \emph{Neural Information Processing Systems (NeurIPS)}, 2019.

\bibitem[Tomani and Buettner(2019)]{trustworthiness}
Christian Tomani and Florian Buettner.
\newblock Towards trustworthy predictions from deep neural networks with fast
  adversarial calibration.
\newblock In \emph{AAAI Conference on Artificial Intelligence}, 2019.

\bibitem[Uzunova et~al.(2019)Uzunova, Ehrhardt, Kepp, and
  Handels]{xai-mp-medical}
Hristina Uzunova, Jan Ehrhardt, Timo Kepp, and Heinz Handels.
\newblock {Interpretable explanations of black box classifiers applied on
  medical images by meaningful perturbations using variational autoencoders}.
\newblock In \emph{Medical Imaging 2019: Image Processing}, 2019.

\bibitem[Vaicenavicius et~al.(2019)Vaicenavicius, Widmann, Andersson, Lindsten,
  Roll, and Sch\"{o}n]{calibration-brier-nll}
Juozas Vaicenavicius, David Widmann, Carl Andersson, Fredrik Lindsten, Jacob
  Roll, and Thomas Sch\"{o}n.
\newblock Evaluating model calibration in classification.
\newblock In \emph{International Conference on Artificial Intelligence and
  Statistics (AISTATS)}, 2019.

\bibitem[{van der Velden} et~al.(2022){van der Velden}, Kuijf, Gilhuijs, and
  Viergever]{xai-medical-survey}
Bas~H.M. {van der Velden}, Hugo~J. Kuijf, Kenneth~G.A. Gilhuijs, and Max~A.
  Viergever.
\newblock Explainable artificial intelligence (xai) in deep learning-based
  medical image analysis.
\newblock \emph{Medical Image Analysis}, 2022.

\bibitem[Veeling et~al.(2018)Veeling, Linmans, Winkens, Cohen, and
  Welling]{dataset-histopathology3}
Bastiaan~S. Veeling, Jasper Linmans, Jim Winkens, Taco Cohen, and Max Welling.
\newblock Rotation equivariant cnns for digital pathology.
\newblock In Alejandro~F. Frangi, Julia~A. Schnabel, Christos Davatzikos,
  Carlos Alberola-L{\'o}pez, and Gabor Fichtinger, editors, \emph{Medical Image
  Computing and Computer Assisted Intervention (MICCAI)}, 2018.

\bibitem[Wang et~al.(2023)Wang, Gong, and Wang]{calibration-study}
Dongdong Wang, Boqing Gong, and Liqiang Wang.
\newblock On calibrating semantic segmentation models: Analyses and an
  algorithm.
\newblock In \emph{IEEE Conference on Computer Vision and Pattern Recognition
  (CVPR)}, 2023.

\bibitem[Wang et~al.(2019)Wang, Li, Aertsen, Deprest, Ourselin, and
  Vercauteren]{uncertainty-aleatoric-tta}
Guotai Wang, Wenqi Li, Michael Aertsen, Jan Deprest, Sébastien Ourselin, and
  Tom Vercauteren.
\newblock Aleatoric uncertainty estimation with test-time augmentation for
  medical image segmentation with convolutional neural networks.
\newblock \emph{Neurocomputing}, 2019.

\bibitem[Wen et~al.(2021)Wen, Chen, Deng, and Zhou]{pretraining-medicalimaging}
Yang Wen, Leiting Chen, Yu~Deng, and Chuan Zhou.
\newblock Rethinking pre-training on medical imaging.
\newblock \emph{Journal of Visual Communication and Image Representation},
  2021.

\bibitem[Zagoruyko and Komodakis(2016)]{wideresnet}
Sergey Zagoruyko and Nikos Komodakis.
\newblock Wide residual networks.
\newblock In \emph{British Machine Vision Conference (BMVC)}, 2016.

\bibitem[Zhang et~al.(2018)Zhang, Ciss{\'{e}}, Dauphin, and Lopez{-}Paz]{mixup}
Hongyi Zhang, Moustapha Ciss{\'{e}}, Yann~N. Dauphin, and David Lopez{-}Paz.
\newblock mixup: Beyond empirical risk minimization.
\newblock In \emph{International Conference on Learning Representations
  (ICLR)}, 2018.

\bibitem[Zintgraf et~al.(2017)Zintgraf, Cohen, Adel, and Welling]{xai-pda}
Luisa~M. Zintgraf, Taco~S. Cohen, Tameem Adel, and Max Welling.
\newblock Visualizing deep neural network decisions: Prediction difference
  analysis.
\newblock In \emph{International Conference on Learning Representations
  (ICLR)}, 2017.

\end{thebibliography}

\clearpage
\onecolumn
\section{Supplementary Material}
\begin{figure}[ht] 
    % \centering
    \centerline{    \includegraphics[width=.65\linewidth,keepaspectratio]{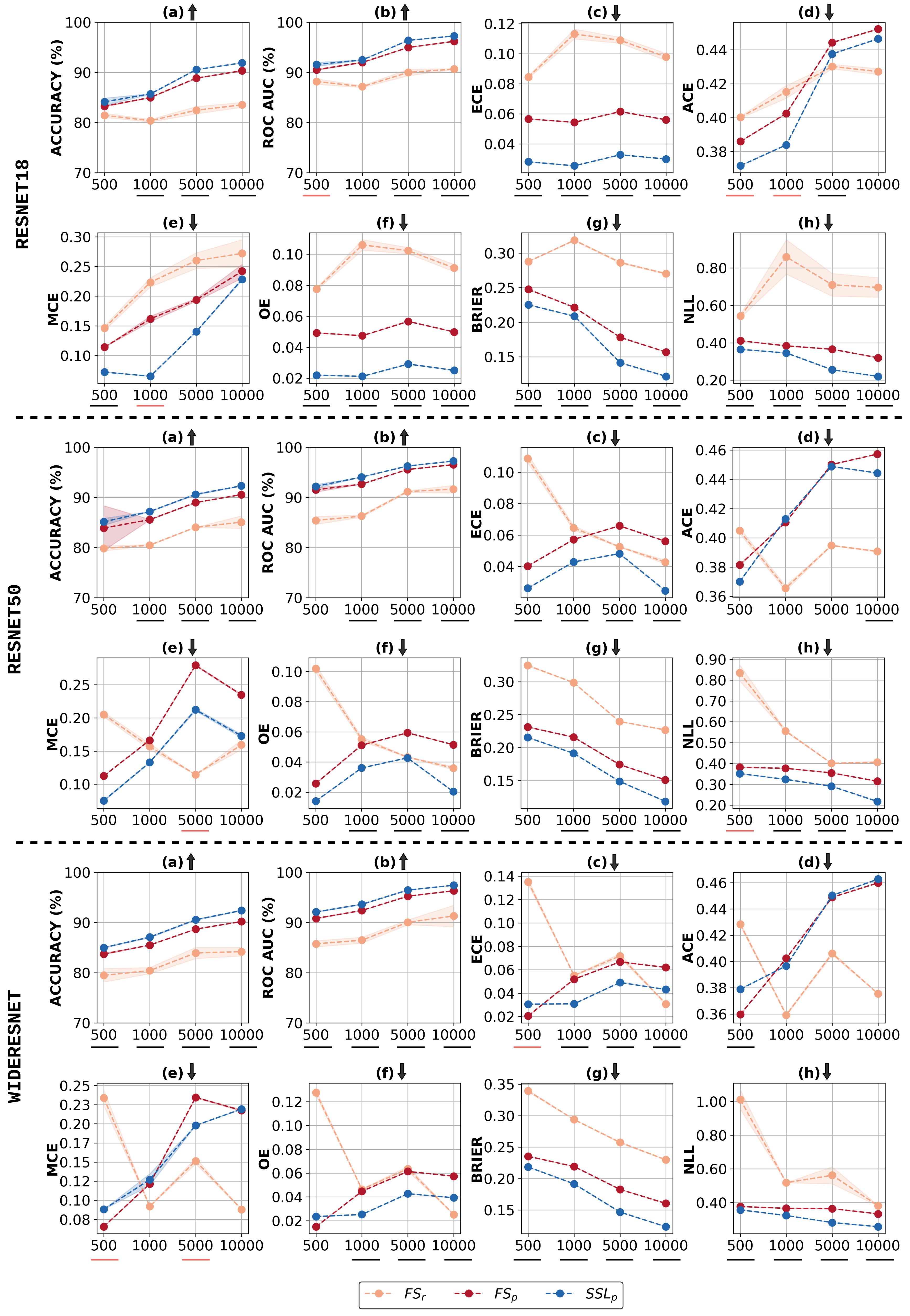}}
    \caption{Joint evaluation for performance and calibration across different dataset sizes (x-axis) and architectures for Histopathology Cancer dataset. The shaded region corresponds to $\mu \pm \sigma$, estimated over 3 trials. The underline shows the statistical significance between $FS_p$ and $SSL_p$. Black and Pink color signifies $p<0.05$ and $0.05<p<0.1$ level of significance, respectively.}
    \label{fig:histopathology_datasizes_and_arch}
\end{figure}

\begin{figure}[ht]
    % \centering
    \centerline{
    \includegraphics[width=.65\linewidth,keepaspectratio]{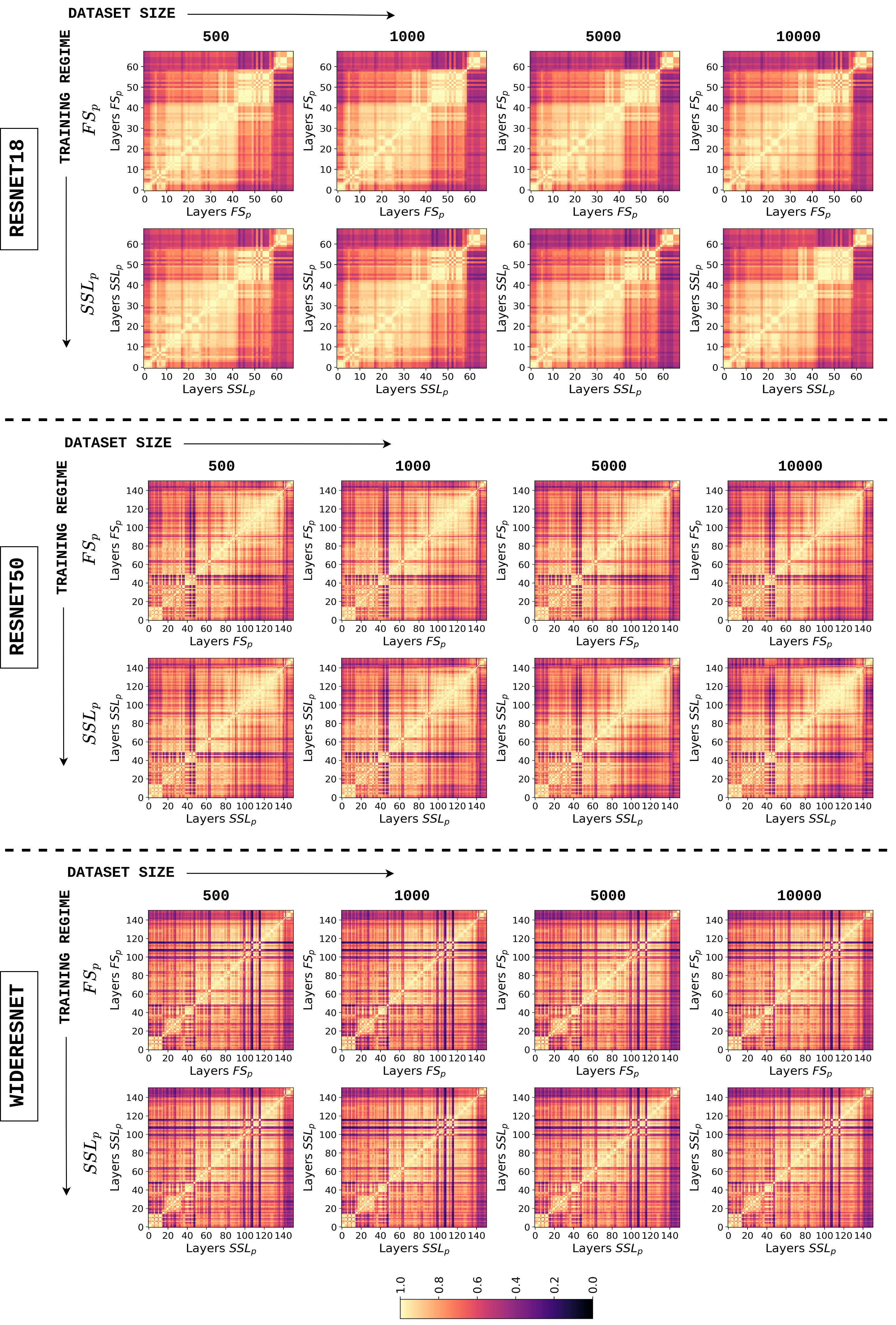}}
    \caption{CKA plots of trained architectures using different regimes for Histopathology Cancer dataset.}
    \label{fig:histopathology_cka_all_arch}
\end{figure}

\noindent The supplementary material is organized as follows: Section 5.1 contains architectures and hyperparameters Details. Section 5.2 investigates the effect of domain-specific transfer learning using RadImageNet pretraining. Section 5.3 consists of the standard deviation of weights distribution vs. calibration scores analysis and CKA plots of ResNet18 and ResNet50 architectures for Diabetic Retinopathy dataset. Section 5.4 contains performance and calibration plots (Figure \ref{fig:histopathology_datasizes_and_arch}) and CKA plots (Figure \ref{fig:histopathology_cka_all_arch}) for Histopathology Cancer dataset. Section 5.5 shows the quantitative results of the CKA analysis using mean CKA values. Section 5.6 compares different training regimes with the reconstruction-based self-supervised task.
\subsection{Architectures and Hyperparameters Details} \label{ssec:sup_hyperparameters}
\subsubsection{Architecture Details}
\noindent We choose three architectures from the ResNet family to evaluate the performance and calibration using three training regimes, $FS_r$, $FS_p$, and $SSL_p$. 
\begin{itemize}
    \item \textbf{WideResNet (\emph{WRN-d-k}:)} It is a variant of residual networks to simulate large architecture size. The depth and width of WideResNet are regulated by a deepening factor \emph{d} and a widening factor \emph{k}. We used \textit{WRN-50-2} for our experiments, i.e., WideResNet with 50 convolutional layers and a widening factor of 2.
    \item \textbf{ResNet50 \& ResNet18:} We choose the standard architectures to simulate medium and small architecture sizes, respectively.
\end{itemize}
\begin{table}[htbp]
  \begin{center}
    \caption{Overview of the models used in this study.}
    \label{tab:table1}
    \begin{tabular}{llr} % <-- Alignments: 1st column left, 2nd middle and 3rd right, with vertical lines in between
      \hline
      \textbf{Model Name} & \textbf{Number of Layers} & \textbf{Parameters}\\
      \hline
    %  ResNet18 11178051
      ResNet18 & 18 layers & 11M\\
    %  ResNet50 23514179
      ResNet50 & 50  layers & 23M\\
    %  Wide ResNet 66840387
      WideResNet & ResNet50, 2$\times$width & 66M\\
      \hline
    \end{tabular}
  \end{center}
\end{table}

\subsubsection{Hyperparameter Details}
\noindent We used the following parameter values for $FS_r$, $FS_p$, and $SSL_p$ training regimes across all datasets and architecture sizes for our experiments. We set batch size=$16$, epochs=$300$, optimizer=SGD, learning rate=$0.001$, momentum=$0.9$, and weight decay=$0.0005$.

For pretrained setups, $FS_p$ and $SSL_p$, we trained the classifier and auxiliary module for the first 30 epochs with a learning rate=$0.001$ and then fine-tuned the complete network with the learning rate=$0.00001$. In $SSL_p$ training, $\lambda \in \{0.1, 0.3, 0.5, 0.7, 0.9, 1.0\} $ is empirically chosen for different datasets and architectures sizes based on the best validation accuracy.

% ##############################
% Diabetic Retinopathy Dataset
% ##############################
\clearpage
\subsection{RadImageNet Pretraining}
\begin{figure}[h] 
    % \centering
    \centerline{
    \includegraphics[width=.7\linewidth,keepaspectratio]{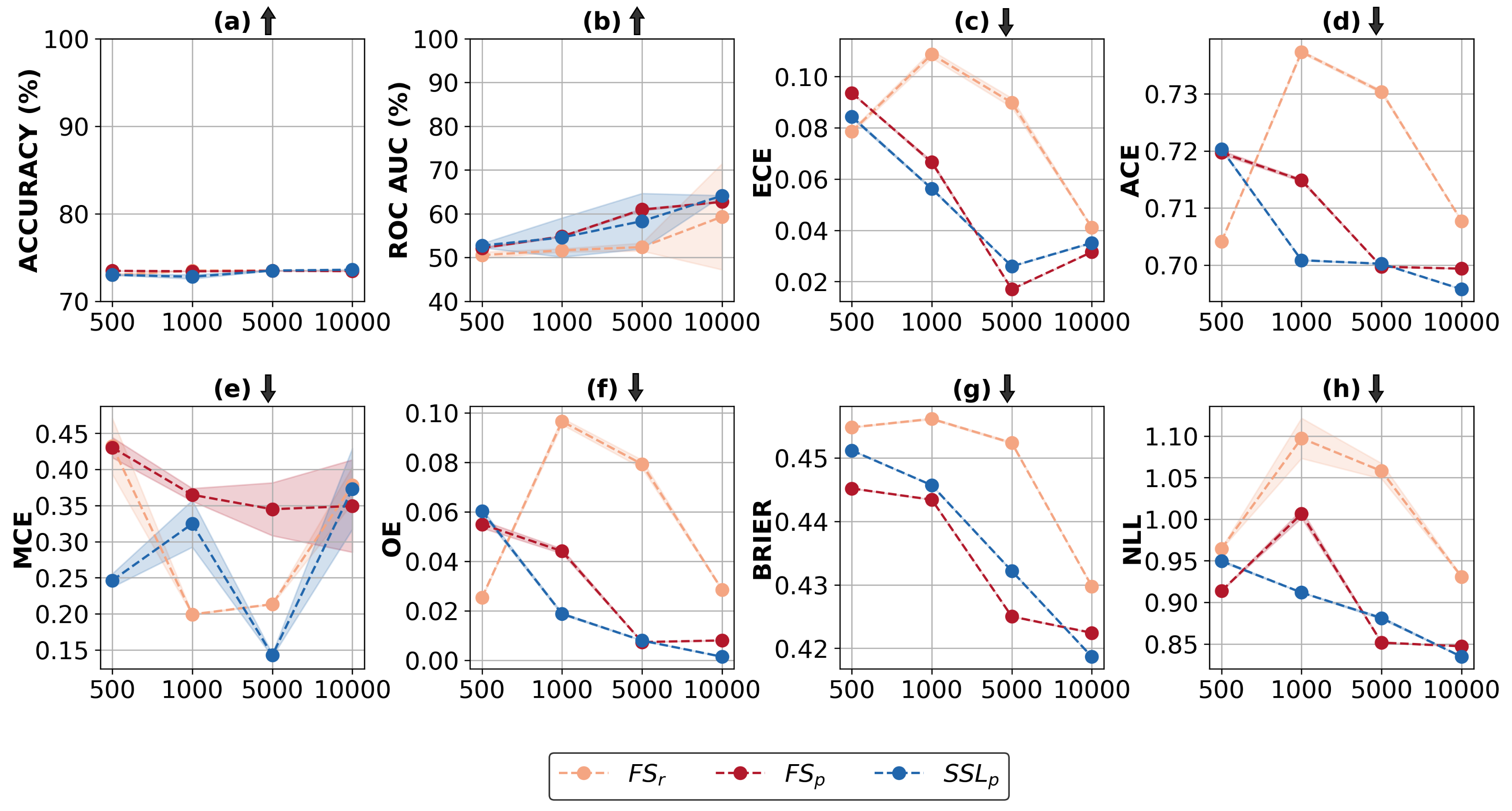}}
    \caption{Joint evaluation for performance and calibration across different dataset sizes (x-axis) of DR dataset using ResNet50 architecture with RadImageNet pretraining. The shaded region corresponds to $\mu \pm \sigma$, estimated over 3 trials. $\uparrow$: higher is better, $\downarrow$: lower is better.}
    \label{fig:dr_radimagenet}
\end{figure}

\noindent To investigate the effect of domain-specific transfer learning, we conducted experiments using RadImageNet {\cite{radimagenet}} \textendash{} a pretrained neural network (ResNet50) trained only on medical imaging datasets shown in Figure {\ref{fig:dr_radimagenet}}. Overall, we notice consistent patterns in calibration, where $SSL_p$ either outperforms or matches $FS_p$, in line with our observations from other experiments. In this context, we observe that $FS_p$ and $SSL_p$ exhibit comparable performance in (a) and (b). However, in the MCE plot (e), $SSL_p$ demonstrates superior calibration compared to $FS_p$. For the remaining metrics, $SSL_p$ tends to show marginal improvement or comparable calibration. Taken together, these findings provide additional evidence that $SSL_p$ consistently delivers calibration models on par with, or sometimes even superior to, those produced by $FS_p$.

\clearpage
\subsection{Diabetic Retinopathy Dataset}
\label{ssec:sup_dr}
\begin{figure}[ht] 
    % \centering
    \centerline{
    \includegraphics[width=.7\linewidth,keepaspectratio]{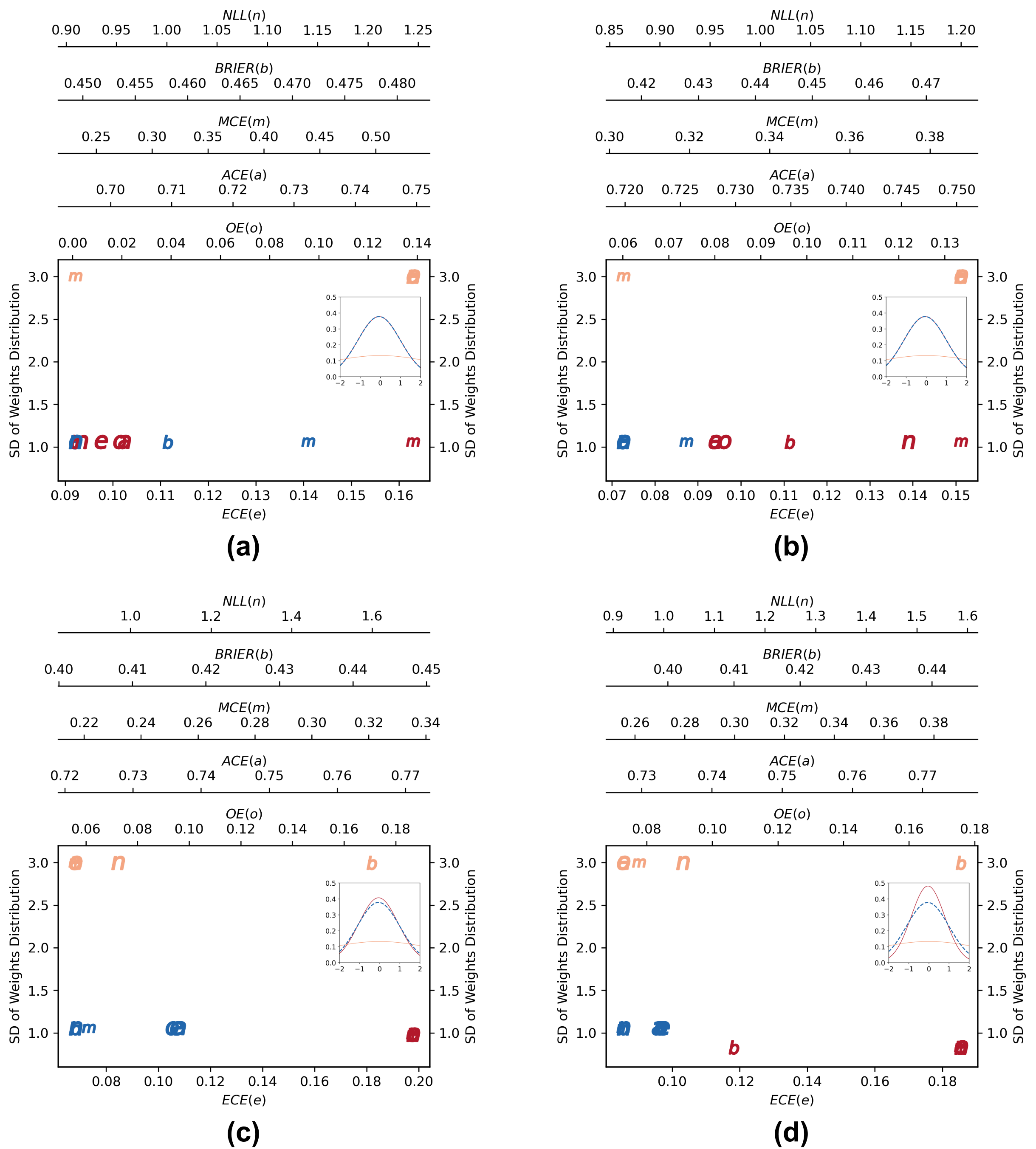}}
    \caption{Standard Deviation of Weights distribution vs. Calibration scores analysis. (a), (b), (c), and (d) depict the relationship between the SD of weights distribution and calibration metrics from the smallest dataset size to the largest one (500, 1000, 1000, 10000), respectively of the DR dataset. Additionally, the corresponding weight distribution plots have been overlaid for convenience of reference. Considering the four plots, we can observe the trend that the calibration metrics of different regimes are segregated when there is a difference in the spread of their distributions (as shown in plots c \& d) and overlapping when there is no difference in the SD of weights distribution (as shown in plots a \& b). Based on the characteristics of $SSL_p$ (shown in blue), it can be remarked that a balance in the spread of weights is necessary to achieve both good performance and calibration.}
    \label{fig:SDvsmetrics_trends}
\end{figure}
\begin{figure}[!h] 
    % \centering
    \centerline{
    \includegraphics[width=.88\linewidth,keepaspectratio]{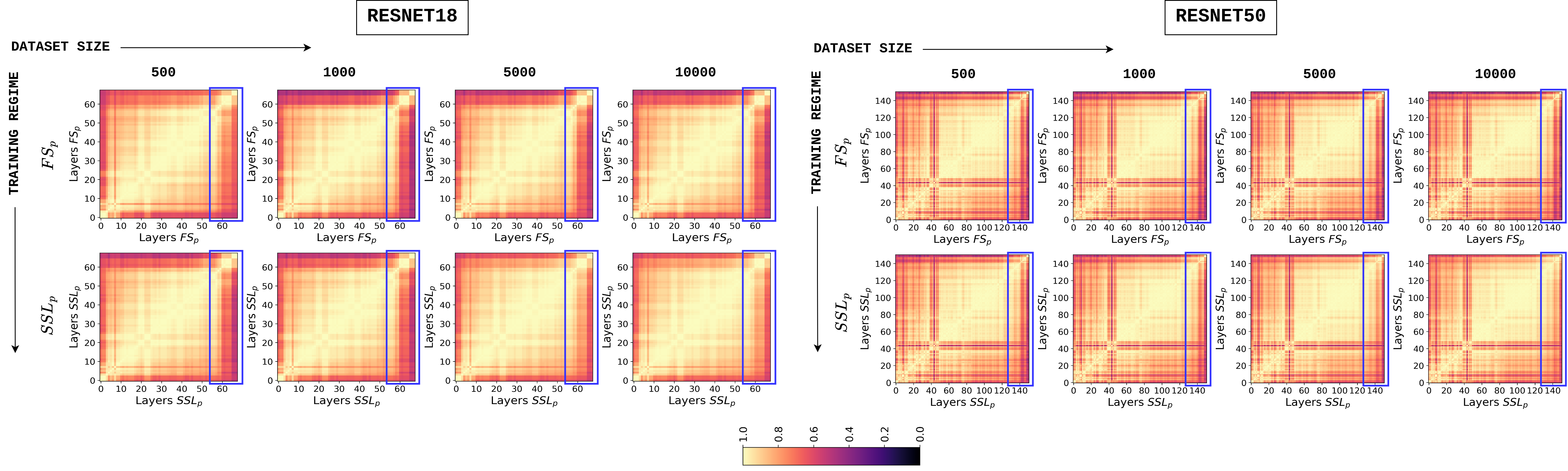}}
    \caption{CKA plots of trained ResNet18 and ResNet50 architectures using $FS_r$, $FS_p$, and $SSL_p$ regimes for DR dataset.}
    \label{fig:dr_cka_resnet18_resnet50}
\end{figure}

\clearpage
% ##############################
% Histopathology Dataset
% ##############################
\subsection{Histopathology Cancer Dataset}
\label{ssec:sup_histopathology}

\subsection{Quantitative Comparison CKA}
\label{subsec:cka_comparision}
\noindent Table {\ref{tab:mean_cka}} presents the quantitative results of the CKA analysis, using mean CKA values. These findings align with the trends observed in Figure {\ref{fig:dr_cka_wideresnet}}. In the case of the DR dataset, the mean CKA values of $SSL_p$ rise as the dataset size increases. This supports our previous findings, where the calibration of $SSL_p$ is superior to that of $FS_p$, and this distinction grows more pronounced as the dataset size becomes larger (Figure {\ref{subfig:SDvsmetrics_dr}}). In the context of the Histopathology dataset, previous observations also indicated that $SSL_p$ outperforms $FS_p$ in terms of calibration, although the difference in calibration metrics values' magnitude is less ({\ref{subfig:SDvsmetrics_histopathology}}). Consequently, we notice that there is no significant difference in the mean CKA values between the two training approaches indicating the representations learned are quite similar.

\begin{table}[h]
% \rmfamily
% \sffamily
\ttfamily
\centering
\caption{Mean CKA values of different training regimes across varying architectures, datasets and their sizes.}
\label{tab:mean_cka}
% \resizebox{.9\linewidth}{!}{ 
\resizebox{.6\linewidth}{!}{ 
  \begin{tabular}{cccccc|cccc}
%   Remove spacing from beginning or end
%   \begin{tabular}{@{}cccccc|cccc|c@{}}
%   \begin{tabular}{@{}lccccc|cccc|c}
%   \begin{tabular}{@{}llllll|llll|l}
    \toprule
    \multirow{2}{*}{\textbf{Architecture}} &
    \multirow{2}{*}{\textbf{\makecell{Training \\ Regime}}} &
    \multicolumn{4}{c|}{\textbf{Diabetic Retinopathy}} & \multicolumn{4}{c}{\textbf{Histopathology Cancer}}\\
    \cmidrule{3-10}
    & & \textbf{500} & \textbf{1000} & \textbf{5000} & \textbf{10000} & \textbf{500} & \textbf{1000} & \textbf{5000} & \textbf{10000}\\ 
    \midrule
    % & $FS_r$ &0.92 &0.91 &0.91 &0.87 &0.90 &0.89 &0.86 &0.80\\
    \multirow{2}{*}{\textbf{ResNet18}} & $FS_p$ &0.86 &0.84 &0.85 &0.85 &0.76 &0.76 &0.75 &0.75\\
    & $SSL_p$ &0.85 &0.85 &0.88 &0.88 &0.76 &0.75 &0.74 &0.75\\
     % \cmidrule{3-11}
    \midrule
    % & $FS_r$ &0.91 &0.85 &0.78 &0.75 &0.74 &0.70 &0.82 &0.66\\
    \multirow{2}{*}{\textbf{ResNet50}} & $FS_p$ &0.84 &0.84 &0.84 &0.84 &0.74 &0.75 &0.74 &0.73\\
    & $SSL_p$ &0.85 &0.85 &0.86 &0.87 &0.75 &0.74 &0.74 &0.73\\
    % \cmidrule{3-11}
    \midrule
    % & $FS_r$  &0.94 &0.82 &0.86 &0.82 &0.84 &0.72 &0.80 &0.75\\
    \multirow{2}{*}{\textbf{WideResNet}} & $FS_p$ &0.84 &0.83 &0.81 &0.81 &0.70 &0.69 &0.69 &0.71\\
    & $SSL_p$ &0.84 &0.85 &0.86 &0.87 &0.69 &0.70 &0.69 &0.71\\
    \bottomrule
  \end{tabular}
  }
\end{table}

\clearpage
\subsection{Comparison of Fully-Supervised and Reconstruction-Based Self-Supervised Task}
\begin{figure}[!h] 
    % \centering
    \centerline{
    \includegraphics[width=.7\linewidth,keepaspectratio]{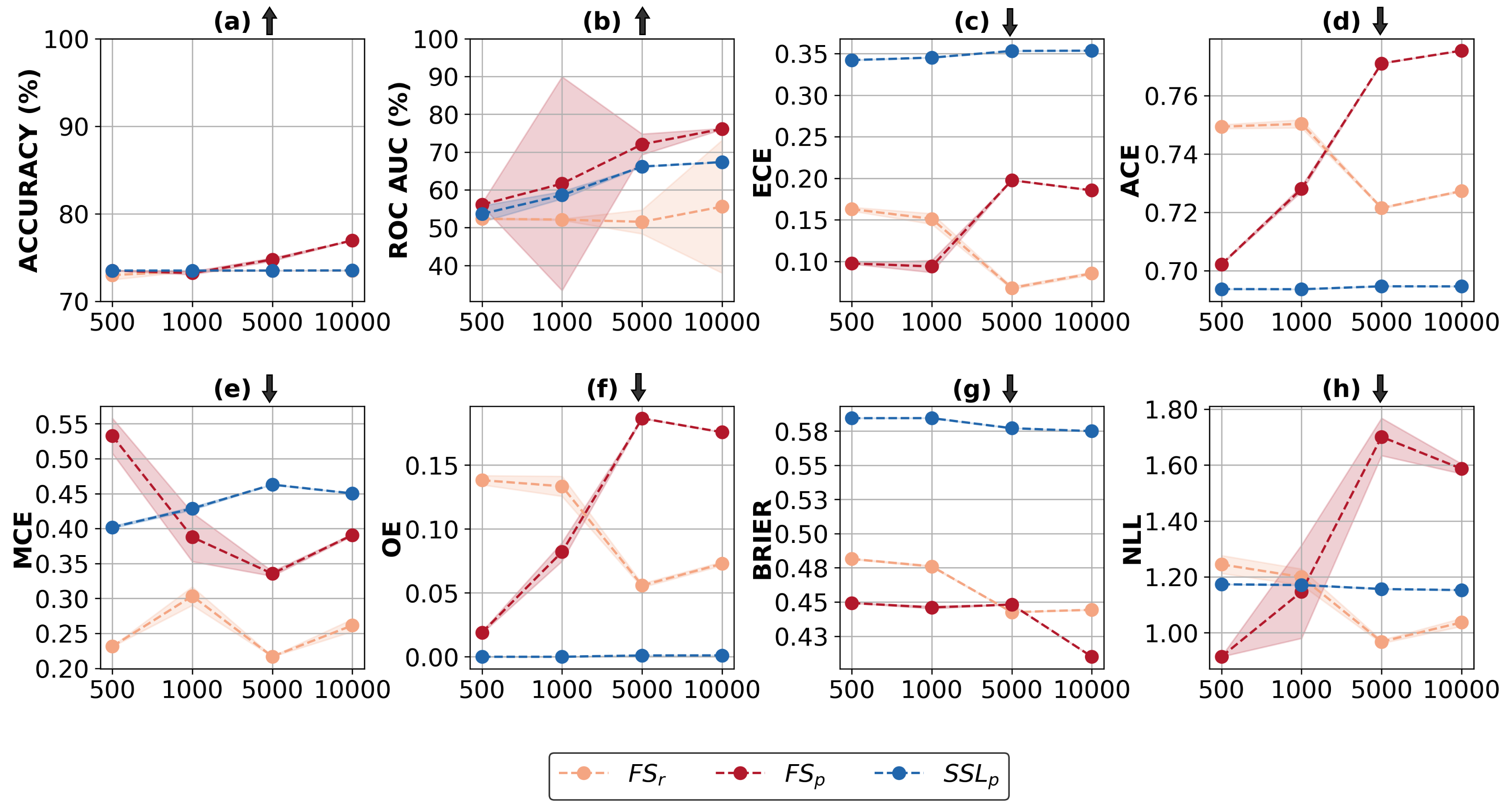}}
    \caption{Comparison of fully supervised ($FS_r$, random initialization), fully supervised ($FS_p$, pretraining), and reconstruction-based auxiliary SSL task ($SSL_p$, pretraining) on DR dataset. Notably, the calibration of models achieved through the auxiliary task does not precisely align with that of the rotation task. Remarkably, the plots reveal a notable contrast: very low OE (f) but high ECE (c). This discrepancy could hint at potential underconfidence, stemming from substantial regularization induced by the reconstruction-based auxiliary SSL task. However, drawing definitive conclusions is premature, as further experiments, encompassing various architectures and hyperparameter tuning, are necessary. Relying solely on the plots, we abstain from making a judgment regarding the superiority of either $FS_p$ or reconstruction-based $SSL_p$.}
    \label{fig:dr_ssl_reconstruction_wideresnet}
\end{figure}

\end{document}